\documentclass[doc,12pt,floatsintext]{apa6}

\usepackage{amsmath}
\usepackage{booktabs}
\usepackage{graphicx}
\usepackage{placeins}         
\usepackage{xcolor}
\usepackage{subcaption}
\usepackage[percent]{overpic}
\renewcommand{\thesubfigure}{\Alph{subfigure}} 
\newcommand{\panellabelfont}{\bfseries} 
\usepackage[bottom,flushmargin, hang]{footmisc}
\usepackage{pslatex}
\usepackage{ragged2e}
\usepackage[natbibapa]{apacite}

\providecommand{\citep}{\citeputhor}

\usepackage{float} 
\usepackage[none]{hyphenat}

\makeatletter
\def\fps@figure{H}
\def\fps@table{H}
\makeatother

\newcommand{\panelimagepos}[3]{%
  \refstepcounter{subfigure}%
  \begin{overpic}[percent,width=\linewidth]{#1}
    \put(#2,#3){\panellabelfont(\thesubfigure)}
  \end{overpic}%
}

\usepackage{etoolbox}
\preto\section{\FloatBarrier} 

\title{{\bf Convolutional Neural Networks Can (Meta-)Learn the Same-Different Relation}}
\shorttitle{(Meta-)Learning the Same-Different Relation}
 
\author{Max Gupta$^1$, Sunayana Rane$^1$, R.~Thomas McCoy$^2$, \\ Thomas L.~Griffiths$^{1,3}$}
\affiliation{$^1$ Department of Computer Science, Princeton University \\ $^2$ Departments of Linguistics and Computer Science, Yale University \\ $^3$ Department of Psychology, Princeton University}

\authornote{\footnotesize
MG acknowledges support from the Princeton AI Teaching Fellowship. We also gratefully acknowledge computational resources from Princeton’s Della HPC cluster. This work was supported by ONR grant N00014-23-1-2510. A preliminary version of these results appeared at the 44th Annual Conference of the Cognitive Science Society \citep{gupta2025convolutional}. \\
\textbf{Correspondence:} Max Gupta, Department of Computer Science, Princeton University, Peretsman Scully Hall, Princeton, NJ 08540. \texttt{mg7411@princeton.edu}.
}

\abstract{While convolutional neural networks (CNNs) have come to match and exceed human performance in many visual reasoning tasks, the tasks these models are optimized for are largely constrained to the level of individual objects, such as classification and captioning. Learning relations -- such as whether two shapes are the same or different -- has been held up as a significant challenge for this class of neural networks.  A number of studies have shown that CNNs tend to generalize poorly to instances of the same-different relation outside of their training distribution, a failure not found in humans or other species that quickly and robustly learn this simple relation. We show that the same CNN architectures that fail to generalize the same-different relation with conventional training are able to succeed when trained via meta-learning, which explicitly encourages abstraction and generalization across tasks. These results provide the first demonstration that the same-different relation can be learned by shallow CNNs trained on relatively small amounts of synthetic data, relevant to considering their value as models of the human visual system.}

\keywords{meta-learning, relational reasoning, convolutional neural networks, abstraction, transfer learning}
\begin{document}

\newcommand{\mccoy}[1]{\textcolor{blue}{\textbf{McCoy: #1}}}
\newcommand{\griffiths}[1]{\textcolor{red}{\textbf{Griffiths: #1}}}

\maketitle

\section{Introduction}

Debates about what aspects of human learning can be captured by artificial neural networks have played a prominent role in the history of cognitive science \citep{minsky1969introduction, rumelhart1986learning, pinker1988language, fodor1988connectionism}. One recent manifestation of this question has focused on the learning capacities of convolutional neural networks (CNNs), which are widely deployed in computer vision models \citep{krizhevsky2012imagenet,lecun2015deep} and have been used to capture aspects of human behavioral and neural responses in object recognition tasks \citep{yamins2014performance,kubilius2019brain, peterson2018evaluating}. Despite strong performance in learning to represent the features of objects, these models have been shown to have difficulty learning about relations between objects \citep{fleuret2011comparing, stabinger201625, kim2018not, puebla2022can}. 

Understanding  relations is central to the human capacity for abstraction.
Perhaps the most basic of our relational abilities -- and arguably a precursor to more complex abstract reasoning -- is the ability to recognize the same-different relation:
are any two given objects the same or not?
Extensive work in cognitive science dating back to the 1980s \citep{premack1983codes} has shown that this ability develops early on in human childhood \citep{blote1999young}, is associated with the learning of language \citep{lupker2015there}, and extends far and wide in the animal kingdom, from bees to ducklings to chimpanzees \citep{gentner2021learning}. However, simple convolutional neural networks have been shown to have difficulty learning the same-different relation \citep{kim2018not}. Recent work has suggested that more sophisticated versions of CNNs can learn some forms of the same-different relation given a well-structured training regimen and a test set that remains close to the training distribution \citep{puebla2022can}, but they still struggle with true out-of-distribution generalization to stimuli never seen during training. These results have led to the tentative conclusion that CNNs may lack the inductive biases needed to learn abstract relational information. 

\begin{figure}[htb!]
    \centering
    \includegraphics[width=\columnwidth]{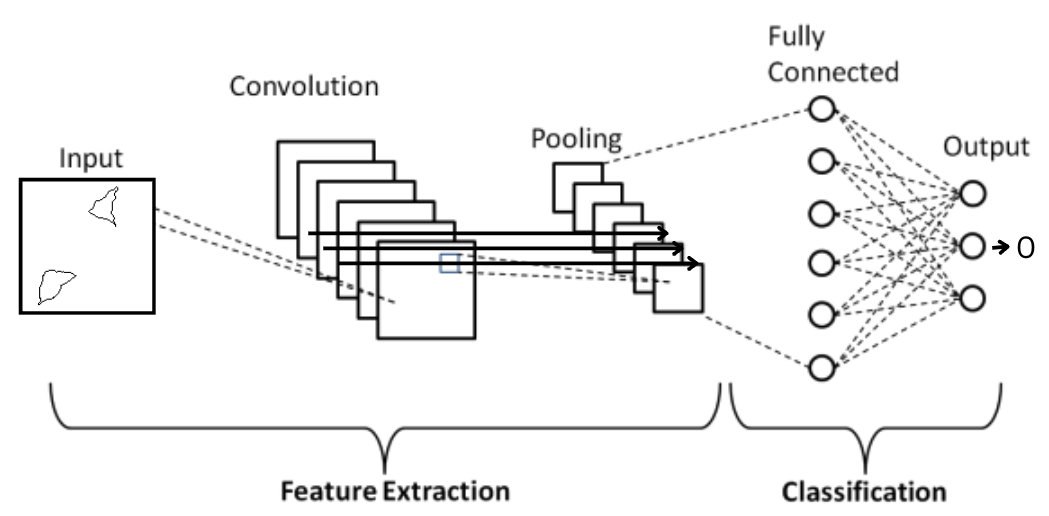}
    \caption{Example of a same-different task as it is posed to a convolutional neural network (CNN) at test-time. Given an image containing two objects, the CNN should return a label of 1 if the two objects are the same or a label of 0 if they are different (as in the example input to the left of the figure). The network processes the image through a series of layers that alternate two operations: convolving learned filters with the pixels in the image to identify features and spatially pooling the results of this computation to induce translation invariance in those features.}
    \label{fig:cnn}
\end{figure}

Previous attempts to train CNNs on the same-different relation have used standard techniques for training neural networks, in which the weights of the network are initialized to random values and then optimized to perform a single task. However, machine learning researchers have made important advances \citep{finn2017model,antoniou2018how} in a different technique for training neural networks, known as meta-learning \citep{schmidhuber1987evolutionary}. Using this technique, a set of neural networks are each trained to perform a different task, but the weights of those networks are all initialized to the same values. The shared initial weights are then optimized to increase the performance of all networks across all tasks. The resulting initial weights encode
the shared structure of the different tasks, making it easier for the individual networks to learn to perform those tasks. Meta-learning has been shown to allow simple neural networks to quickly learn to perform tasks that previously were assumed to require symbolic representations, such as learning formal languages \citep{mccoy2025modeling} or logical concepts \citep{marinescu2024distilling}. In this paper, we explore whether this tendency to find generalizable abstractions is sufficient to allow CNNs to learn the same-different relation and generalize robustly to novel, out-of-distribution stimuli.

We find that meta-learning enables convolutional neural networks to robustly acquire the same-different relation, despite failing to do so with conventional training. In-distribution, simple CNNs trained via meta-learning achieve near-perfect performance across a wide range of shape-based tasks, while conventional training enables only chance performance on most tasks. More strikingly, in out-of-distribution tests where entire shape families are withheld during training, meta-learning allows these CNNs to generalize reliably, surpassing the performance of much larger and more sophisticated architectures trained with conventional methods. These advantages extend to naturalistic stimuli as well, where meta-learning permits CNNs to generalize across novel object categories far more effectively than conventional training. Together, these results demonstrate that a shift in the training paradigm -- optimizing networks for abstraction across tasks -- can allow neural networks to rapidly learn relations previously thought to require symbolic representations. At a high level, our findings suggest that meta-learning provides a powerful route for building domain-general learners that achieve human-like flexibility in acquiring abstract relations.

\section{Background}

\subsection{Convolutional neural networks}

Convolutional neural networks are a type of multi-layered artificial neural network that takes pixel-level visual data as input \citep{lecun2015deep}. Key components of the CNN architecture take  inspiration from biological visual processing systems \citep{hubel1959receptive}. The initial layers of the network learn filters that are applied across an image, with their outputs being spatially pooled to form representations that are translation invariant and expressed at different scales (Figure 1). The learned filters have been shown to detect features, such as edges, which are useful for tasks such as image classification. One important property of a CNN is the number of processing layers that it contains; this number is referred to as its depth. Modern CNNs use architectural advancements such as residual layers \citep{he2016deep} to train deeply layered models on a variety of tasks. CNNs first came to prominence for their remarkable image classification ability \citep{krizhevsky2012imagenet}, and have since matched or surpassed human performance on a variety of visual tasks \citep{alzubaidi2021review}. This has encouraged the use of CNNs as models for human visual processing \citep{yamins2014performance,kubilius2019brain, peterson2018evaluating}. 

\subsection{Learning the same-different relation}

Various forms of CNNs have previously been tested on relational visual tasks. A common dataset used for training and evaluation is the Synthetic Visual Reasoning Test (SVRT) dataset, a battery of 23 different visual-relation tasks \citep{fleuret2011comparing}. In early experiments, CNN architectures that were very successful in image classification tasks were largely unsuccessful on the visual-relation tasks in the SVRT dataset \citep{stabinger201625}. In particular, while CNNs were able to learn spatial relations in this dataset, they performed poorly on relations that involved evaluating whether two shapes were the same \citep{kim2018not}. These findings suggested that CNNs might lack the  inductive biases necessary for learning the same-different relation. 

As CNN architectures improved and showed heightened performance on computer vision problems, further studies investigated whether these more sophisticated architectures were better at learning relations. A study using a CNN architecture with increased multi-layered attention mechanisms, for example, showed significantly improved performance on a range of relational classification tasks \citep{wang2016relation}. This suggested that adding in significant architectural inductive bias could improve performance on relational tasks. However, further studies yielded mixed results. \citet{puebla2022can} applied various CNN architectures to same-different tasks and found that while the networks could perform well on tasks that were similar to the tasks in their training data, their performance dropped significantly when tested on another family of same-different tasks that were superficially different from those in the training data. This finding also held for larger CNNs using the ResNet architecture \citep{he2016deep}, in which residual links from earlier layers of the network to later layers make it possible to train larger and deeper networks. These results supported the conclusion that abstract same-different relations were difficult for CNNs to learn in a generalizable manner. 

Despite these negative results, recent work has suggested that it may be possible for neural networks to represent the same-different relation. Vision-transformer models \citep{dosovitskiy2020image} pre-trained on the large ImageNet dataset show a sensitivity to the same-different relation \citep{tartaglini2023deep}. The observation that some large neural networks are able to learn this relation given sufficient training motivates re-investigating  whether CNNs have the same capacity: if it is possible for neural networks to represent the same-different relation, it may just be a matter of finding an effective training regime that allows them to learn it.


\subsection{Meta-learning}
While previous work has suggested that CNNs may lack the architectural inductive biases needed to robustly learn abstract visual relations such as same-different, this work has been done within a single training paradigm. All evaluations of CNNs in previous work have learned a set of weights by training models to perform a single task (or a set of tasks that are combined into what is effectively a single task). In this paper we use a different approach: meta-learning. In particular, we focus on the Model-Agnostic Meta-Learning (MAML) algorithm \citep{finn2017model}, which is designed to find the optimal initial weights for learning to perform a set of related tasks, such that the model can rapidly generalize to new, unseen tasks. 

\begin{figure}[t]
    \centering
    \includegraphics[width=0.8\columnwidth]{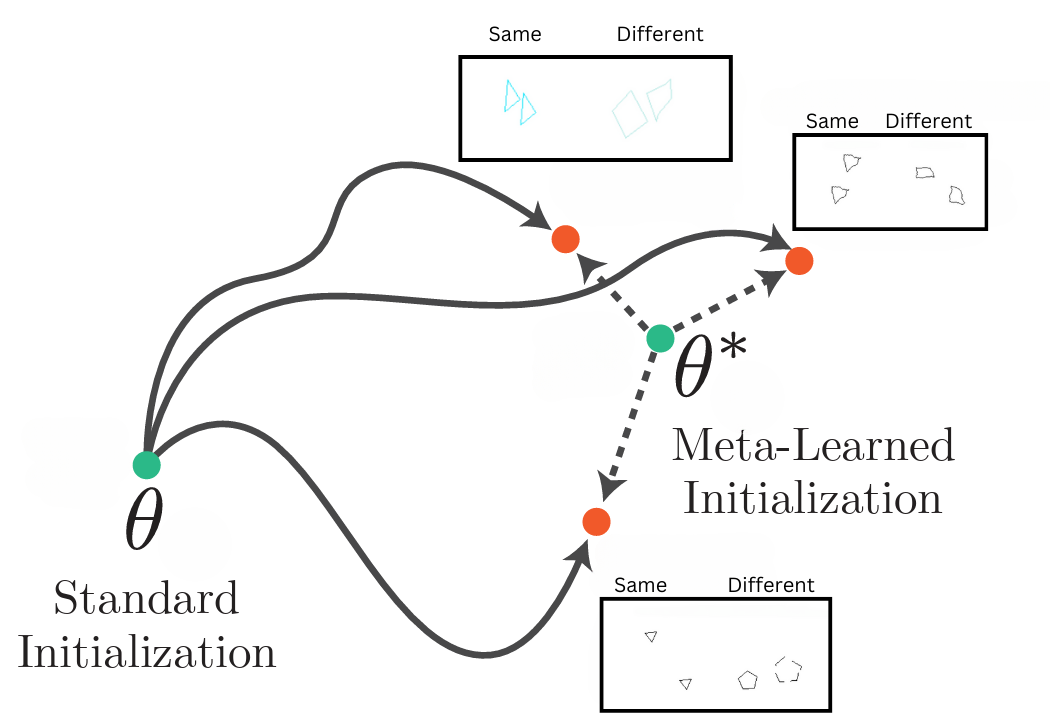}
    \caption{Meta-learning initial weights for generalization. A neural network with a standard initialization ($\theta$) typically requires a large amount of training to learn a specific task (gradient descent needs to follow a long trajectory, shown with the solid arrows). Meta-learning optimizes the network's initialization to create a meta-learned initialization $\theta^*$ from which a range of different tasks can be learned with a small amount of training (represented by the shorter dashed arrows). Here, three examples of new ``same-different'' tasks are shown, learned from the meta-learned initialization over fewer gradient steps than from the standard initialization.}
    \label{fig:metaexplanatory}
\end{figure}

Given a set of tasks ${\cal T}$ such that each task $t \in {\cal T}$ has an associated loss function ${L}_t$, conventional neural network training seeks a set of weights for a neural network  $\phi$ that minimizes the loss function 
\begin{equation}
{\cal L}_{\rm \, conventional} = \sum_{t \in {\cal T}} { L}_t (\phi) 
\end{equation}
which is simply the sum of the losses across different tasks. By contrast, MAML seeks to find the initial weights $\theta$ that minimize the loss function
\begin{equation}
{\cal L}_{\rm \, MAML} = \sum_{t \in {\cal T}} { L}_t ( \phi_t) \qquad {\rm for} \ \phi_t = \theta - \alpha \nabla { L}_t(\theta) 
\end{equation}
where $\phi_t$ are a set of weights adapted for performing task $t$ via gradient descent applied to the loss $L_t$ of task $t$ starting at the initial weights $\theta$ (with $\alpha$ being a learning rate). The resulting $\theta^*$ should capture the regularities shared by the tasks in ${\cal T}$, supporting fast adaptation to unseen, new tasks (see Figure~\ref{fig:metaexplanatory} and Appendix B for more details on the algorithm).

Meta-learning has previously been shown to be effective as a tool for transferring symbolic inductive biases to  neural networks \citep{mccoy2025modeling,marinescu2024distilling} and for producing neural networks that are capable of compositional generalization \citep{lake2023human}. One factor driving these successes in capturing symbolic behavior is the fact that meta-learning focuses on identifying the commonalities across tasks, which in itself is a form of abstraction. In this work, we explore whether meta-learning allows convolutional neural networks to form generalizable representations of the same-different relation. We replicate previous studies testing CNNs of varying depths on these tasks and switch the conventional training regime to one based on meta-learning, looking for the emergence of a reliable generalization of same-different understanding across novel stimuli and in novel tasks. Importantly, the networks that we train are exposed to exactly the same training data regardless of whether the networks are optimized using conventional training or meta-learning; the only difference is the type of optimization algorithm that is applied to the data. Matching the data allows us to isolate the effect of this algorithm on the learning behavior of the model classes we test. 

\section{Replicating Previous Work}

\begin{figure}[bt]
    \centering
    \includegraphics[width=\linewidth]{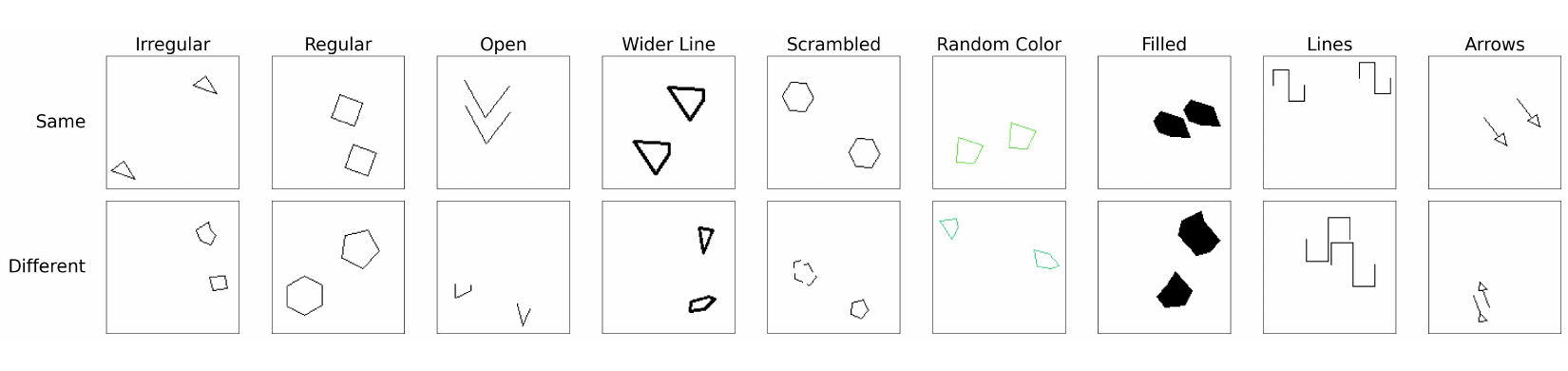}
    \caption{The Same-Different dataset from Puebla and Bowers (2022). Each column shows one of the nine tasks in this dataset, where all tasks are based around the same-different relation but use different types of shapes to instantiate that relation. Each task has a stochastic function generator that ensures each example is unique within any given dataset. }
    \label{fig:augmented}
\end{figure}

All the experiments presented in this paper use the same CNN architectures evaluated by \citet{kim2018not}, who showed that CNNs performed poorly on the same-different relations in the SVRT dataset. This includes CNNs of varying depths and convolutional filter sizes (the exact hyper-parameters specified in the original paper are shown in Appendix A). In addition to the original Problem 1 from the SVRT challenge (pictured in Figure~\ref{fig:cnn}) we also train on nine same-different tasks created by \cite{puebla2022can}, which augment the standard SVRT dataset with new shapes such as arrows, irregular polygons, and shapes with random colors (the full set of tasks we train on is shown in Figure~\ref{fig:augmented}). In the conventional learning setting, same-different tasks are processed as individual input/label pairs: an image and a corresponding binary label (0 or 1 for different or same; Figure~\ref{fig:cnn} illustrates this at test-time, where the label is withheld). Note that, in these datasets, what constitutes a distinct ``task'' is a particular type of shape over which same-different judgments are made. For instance, one task is based on irregular polygons while another is based on regular polygons. Thus, all tasks target the same abstract relation (same-different), but they instantiate this relation with different types of shapes. An example of a full episode as presented to the models is shown in Appendix A.

To establish a baseline, we first evaluate the performance of three CNN architectures in this setting, training each model end-to-end on a dataset containing examples from all 10 tasks, with equal frequency for each task and for the categories of \textit{same} and \textit{different}. We test the CNN architectures used by Kim et al., with 2, 4, and 6 convolutional layers using max pooling, batch normalization, and ReLU activation, followed by 3 fully connected layers of 1024 units each and a 2-dimensional classification layer. All models are trained with Adam optimization \citep{kingma2015adam} and a base learning rate of 1e-3. 

\begin{figure}[htb!]
    \centering
    \includegraphics[width=0.85\textwidth]{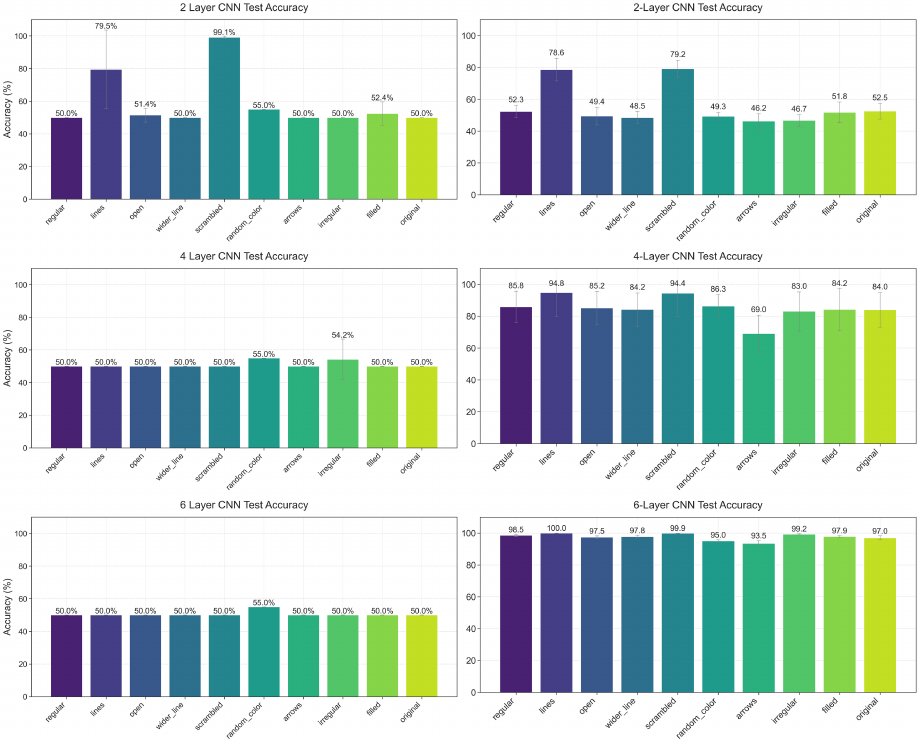}
    \caption{In-distribution same-different classification accuracy for a conventional learner (left) versus a meta-learner (right) by task and architecture. The conventional learner is trained using standard gradient descent, while the meta-learner is trained with the MAML algorithm. Each bar is one version of the same-different task, where the task versions differ in terms of what types of shapes are used to illustrate same-different relations.  ``Original'' represents task \#1 from the SVRT dataset. Error bars represent standard deviations from mean accuracy across 10 randomly initialized seeds.}
    \label{fig:fig4-1}
\end{figure}

We then test each model on unseen same-different examples from the 10 tasks the model has been trained on, as a basic in-distribution test of learning these tasks. Averaging over 10 re-runs of the model with different random seeds, each run to convergence, we find performance stabilizing almost exactly at the level of random guessing for all three model depths (Figure~\ref{fig:fig4-1}, left); since there are two possible labels (\textit{same} and \textit{different}), random guessing would yield an accuracy of 50\%. As an exception to this general trend, some conditions do manage to reliably converge to high-accuracy solutions (in particular, a 2-layer CNN achieves 99\% test-time accuracy on the scrambled task and 80\% accuracy on the lines task). A potential explanation is that the scrambled and lines tasks are the only tasks in the dataset that feature only straight, right-angled lines, a low-level feature that develops in earlier layers of CNNs \citep{krizhevsky2012imagenet}. 

\section{Meta-learning Same-Different In-Distribution}

Having replicated the chance-level performance of this class of CNNs on the augmented same-different dataset, we move to formalizing the meta-learning setup we will use. Without changing the model architecture or the content or quantity of data, we change the learning algorithm from conventional stochastic gradient descent to MAML \citep{finn2017model,antoniou2018how} in order to explore the impact of using meta-learning.  

In the meta-learning setting, we generate ``episodes'' unique to each of the tasks in Figure 3. Each episode consists of labeled example images for task-specific adaptation and query images with held-out labels for evaluation. The meta-learner then has a chance to ``practice'' on the episode's training examples before outputting predictions for each query image. Each episode contains different, randomly-generated examples from exactly one task. Within-episode learning constitutes the ``inner loop'' of conventional learning  of a single task -- taking a gradient step from the initial weights $\theta$ -- for which we provide a ``fast'' inner learning rate of 1e-3. To aid the model in learning from a diversity of episode structures, we randomly intersperse tasks and use variable numbers of evenly distributed same/different examples (4, 6, 8, and 10 examples; see Appendix A) and fixed query sizes (2 examples, always). Training details and hyperparameter values are listed in Appendix A.

The outer loop consists of optimizing the initial weights $\theta$ used in the inner loop. This involves the more gradual transfer of knowledge from one task to other tasks, for which we assign a ``slow'' learning rate of 1e-4 and use Adam optimization. Following standard methodology, we add a validation stage in which we monitor model performance on unseen episodes from the same set of tasks. This outer-loop update optimizes for generalization across tasks, creating pressure for the learner to go beyond example and task-specific properties toward a more generalizable notion of sameness and difference. This kind of generalization is further encouraged by our method of sampling episodes that alternate across task type and support size.

We ensure that our meta-learning neural networks (described in this section) and our conventional neural networks (described in the previous section) receive exactly the same training data. This is done using the following procedure: We first generate the dataset for the meta-learning setting as described in the previous paragraphs, sampling a set of examples and a set of queries for each of the 1000 episodes that we produce for each of the 10 tasks. The training data used for the conventional procedure summarized in the previous section are then created by ``flattening'' this meta-learning dataset -- that is, the conventional training set is the concatenation of all sets of examples and queries from all episodes in the meta-learning training set. 

As a first step, we test the three previously described CNN architectures in our meta-learning setup by meta-training on 1000 episodes from each of the 10 tasks and testing on unseen episodes from the same tasks. A 2-layer CNN performs at chance level on the majority of tasks seen in-distribution, but we see a striking increase in performance as we increase the depth of the network (Figure 4, right). A CNN with 6 convolutional layers performs at almost perfect accuracy across all tasks it has been meta-trained on, and its accuracy is consistent across re-runs. For full results, see Appendix A. This suggests that these deeper networks are better able to respond to the pressure for abstraction that we intend to create via meta-learning. Deeper networks have more tunable parameters, thus potentially benefiting more from setting those parameters to values that result in good features by finding good initial weights.

A potential reason for this strong performance on data sampled in-distribution (as shown here) could be explained by the network having simply memorized shallow properties of its data distribution, rather than having generalized the abstract notion of sameness and difference ---a concern that motivates us to study more stringent tests of generalization in the next sections.

\section{Meta-learning Same-Different Out-of-Distribution: Leave-One-Out}

Having shown that CNNs can meta-learn the same-different relation when tested on further examples generated from the same distributions as the training tasks, we now assess their ability to generalize this relation by testing on unseen, out-of-distribution tasks. To do so, we perform a leave-one-out test, training the previously highest performing model configuration (the 6-layer CNN) on the same battery of same-different classification problems, but crucially holding out one task from training for testing. That is, each model is trained on nine tasks and generalizes to the remaining task, rotating which task is being held out in this way. This allows us to systematically test the models on an out-of-distribution task they have never seen during training. We do this for all tasks and meta-train to convergence with MAML, using the same parameters described above. 
 
In using this design we intentionally replicate the structure of an experiment conducted by \cite{puebla2022can}, who found that even much larger, pre-trained ResNet architectures were unable to generalize out-of-distribution on this same set of tasks using conventional learning techniques. By using the same design in a meta-learning context, we can directly evaluate the impact of meta-learning on generalizable learning of the same-different relation. 

\begin{figure}[htb!]
    \centering
    \includegraphics[width=0.9\textwidth]{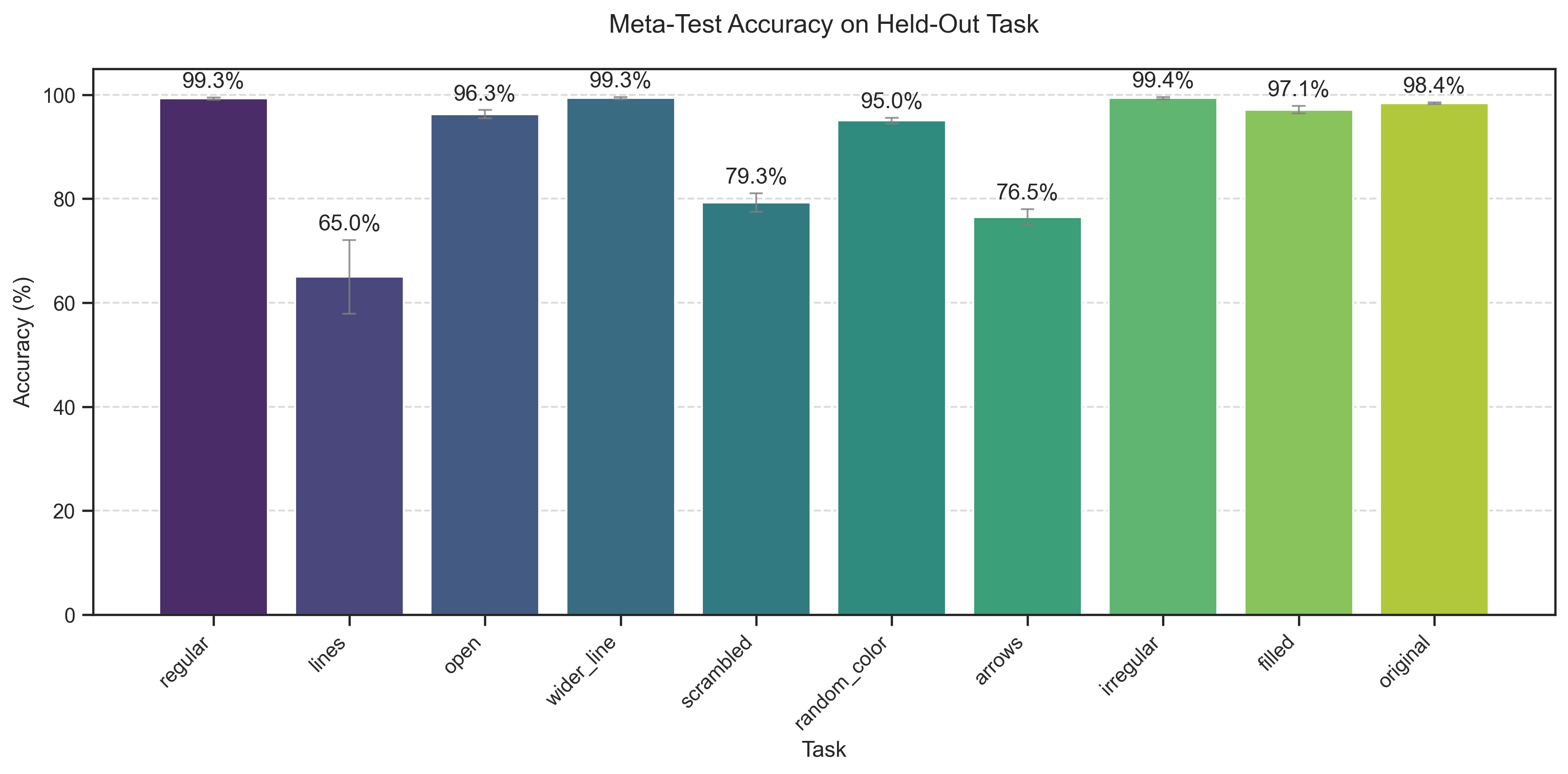}
    \caption{Out-of-distribution same-different classification accuracy for a CNN trained using meta-learning. Each bar shows one version of the same-different task, where the versions differ in terms of the types of shapes used to illustrate same-different relations. A separate CNN was meta-trained to produce the results shown for each bar. The meta-training process included all tasks except the one to be evaluated on, and the resulting network was then trained and evaluated on that withheld task. Thus, the task being evaluated on was always out-of-distribution.}
    \label{fig:fig5}
\end{figure}

Figure~\ref{fig:fig5} presents the results of using this procedure, showing that even this shallow class of CNNs can reliably generalize to the most challenging out-of-distribution tasks when trained via meta-learning. Classification accuracies in our evaluations were about 95\% for seven of the held-out tasks, averaged over 10 seeds (randomly initialized weights), with lower performance in three tasks (arrows, lines, and scrambled). These three tasks were also found to be the hardest for ResNets in the analysis performed by \citet{puebla2022can}. However, meta-learning allows even these shallower CNNs to outperform the results for ResNets reported by \citet{puebla2022can} in each of these ``failure'' cases by significant margins. For example in the worst performing task (``lines''), previous results showed ResNet50 and ResNet152 performing at chance accuracy, whereas we show a meta-trained CNN consistently performs above chance averaged over 10 seeds (where chance level is 50\%). In Appendix C we provide sample efficiency details comparing MAML and conventional training in this setting.\footnote{Although \citet{puebla2022can} did not release exact details on the amount of data used in training, ImageNet pretraining typically allows the models to see over 1 million images prior to fine-tuning on the same-different task.}

\section{Naturalistic Experiments}

The results for same-different generalization shown so far have involved training on stochastically-generated synthetic shapes. Ideally, our models should be able to learn to generalize the same-different relation to novel stimuli from the real world. To push tests of generalization further, we conducted a series of experiments testing our meta-trained CNNs on same-different pairings of real world objects collected from the Google Search and commercial datasets by \citet{brady2008visual}, inspired by the experiments on pre-trained vision transformers presented by \citet{tartaglini2023deep}. We examine two examples of learning transfer: transfer from synthetic data to naturalistic data and transfer from a subset of naturalistic data to a held-out set of naturalistic data. As before, we compare performance across models trained with meta-learning versus conventional learning. 


First, as the most stringent test of far out-of-distribution generalization, we examined whether training on just the synthetic data from \citet{puebla2022can} is sufficient for CNNs to generalize the same-different relation to naturalistic objects (an example is shown in Figure 6). We load the weights of all the models meta-trained in the previous experiments and test them on same-different pairings of naturalistic objects (with no training on tasks involving those naturalistic objects). This far out-of-distribution generalization is challenging for all models, given the significant increase in pixel density and complexity of the naturalistic stimuli (see Figure 6B). We find that the same-different information learned from training on the data from \citet{puebla2022can} only allows the deepest model (6-layer CNN) to generalize reliably above chance to naturalistic stimuli, consistent with our previous experiments in which deeper models appeared to generalize better. 


Next, we wanted to test whether CNNs can generalize to naturalistic stimuli given the chance to train on a subset of those naturalistic stimuli. We compared MAML and conventional training: meta-trained models used 600 epochs of 1,000 episodes each, while conventional models trained for 300 epochs on 6,400 images per epoch. Because conventional learning takes longer to reach ceiling performance, in this way all models were trained to convergence. A subset of stimulus pairings were held-out for evaluation, and performance was measured via classification accuracy on held-out examples that were not presented during training. All methods used the same parameters, detailed in Appendix A. 


The results produced by this procedure were similar to the results from our previous analyses: given the chance to train on just a subset of naturalistic stimuli during meta-training, CNNs are able to generalize out-of-distribution across all model depths, with accuracy increasing with increasing depth (see Figure 6C). However, as was the case with synthetic data, conventionally trained models failed to generalize robustly here as well. 


These results hint at interesting distributional limitations of meta-learning, previously documented in \citet{bencomo2025teasing} and \citet{hasson2020direct}: while meta-learning allows rapid test-time adaptation to unseen tasks, generalization is still constrained to a large extent by the range and dimensionality of the training data. In our case, the stimulus space of the naturalistic data we test on is far richer and more complex pixel-wise than the geometric shapes used in previous studies and thus encodes a far wider and more difficult learning challenge, explaining the more modest results in Figure 6B. However, the fact that such size-limited CNNs can reliably meta-learn the full space of naturalistic stimuli from training on just a sub-sample of data is impressive and points to the domain-general power of meta-learning for relational learning across visually challenging datasets. 

\section{Analyzing the Weights of the Networks}

While we have shown the advantage of meta-learning for test-time performance in these models, we would also like to be able to better mechanistically understand how meta-learning allows CNNs to arrive at these solutions. We aimed to gain such an understanding by analyzing the weights of both meta-trained and conventionally-trained networks. Given the numbers of weights in our models (exceeding 8 million tunable weights), we use Principal Component Analysis (PCA), a dimensionality reduction technique, to effectively project the weights of the best-performing models (6-layer CNNs) into a 2-dimensional space for visualization. Our goal in doing so is to see whether we can recover a version of the diagram in Figure 2 which showed that, theoretically, the weights of our meta-learners should be as close as possible to the optimal weights for all tasks in the family of same-different tasks in Figure 3. 

\begin{figure}[tb!]
  \centering

  \newcommand{\panelimage}[1]{%
    \refstepcounter{subfigure}%
    \begin{overpic}[width=\linewidth]{#1}
      \put(2,92){\panellabelfont(\thesubfigure)}
    \end{overpic}%
  }

  \begin{subfigure}[t]{0.65\textwidth}
    \centering
    \panelimagepos{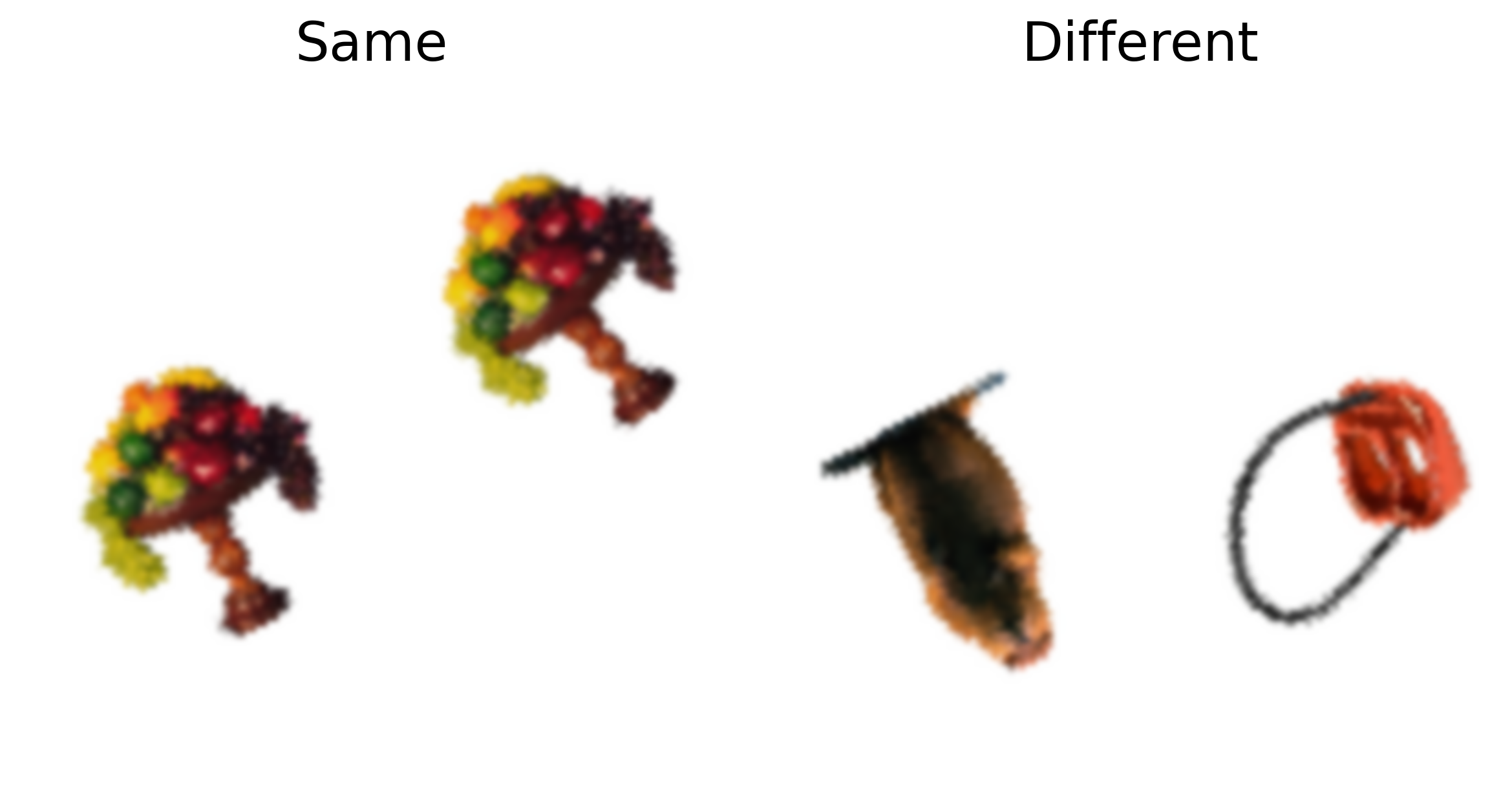}{2}{46}
    \label{fig:panelA}
  \end{subfigure}

  \begin{subfigure}[t]{0.45\textwidth}
    \centering
    \panelimagepos{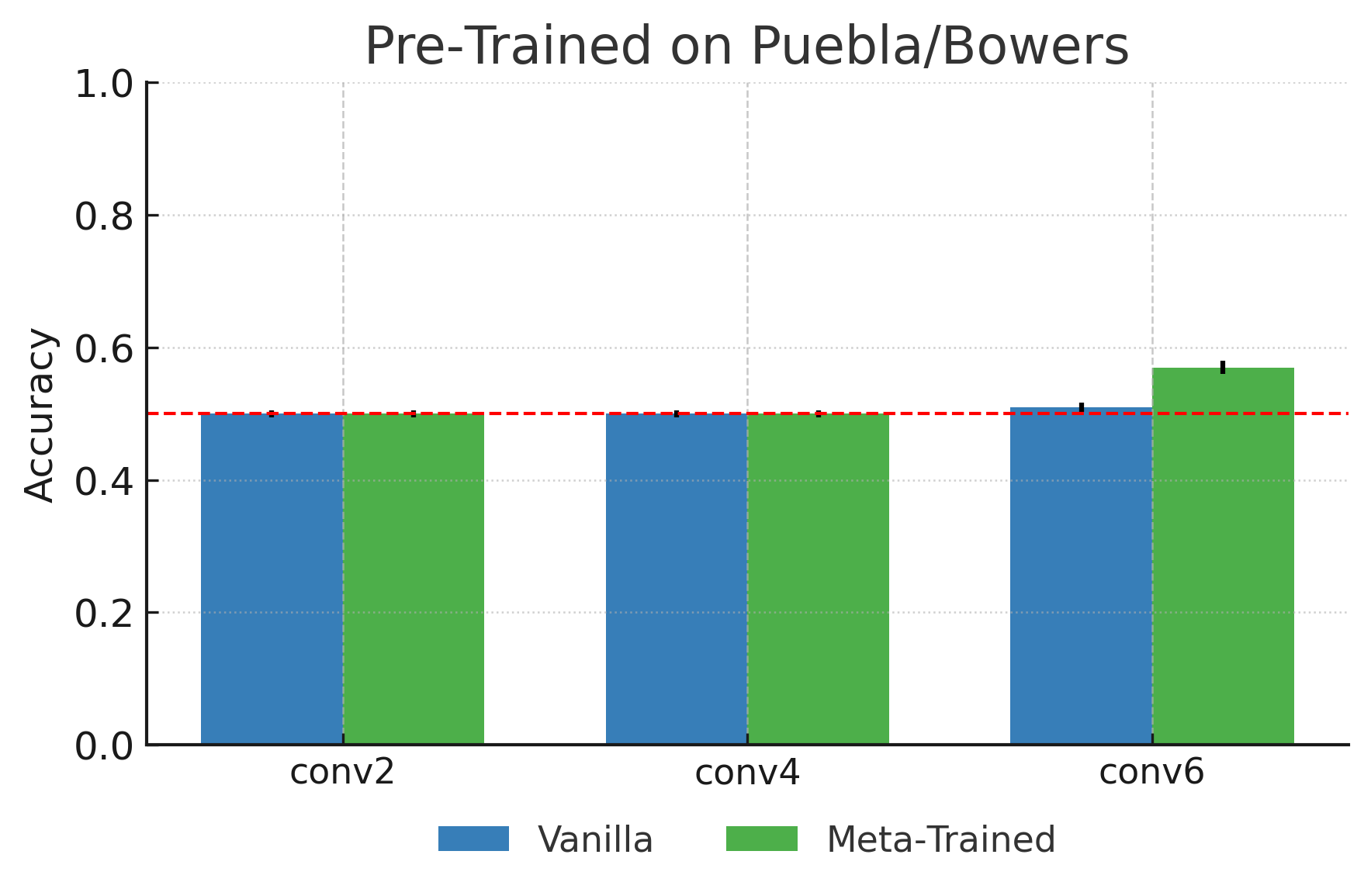}{2}{77}
    \label{fig:panelB}
  \end{subfigure}
  \hfill
  \begin{subfigure}[t]{0.45\textwidth}
    \centering
    \panelimagepos{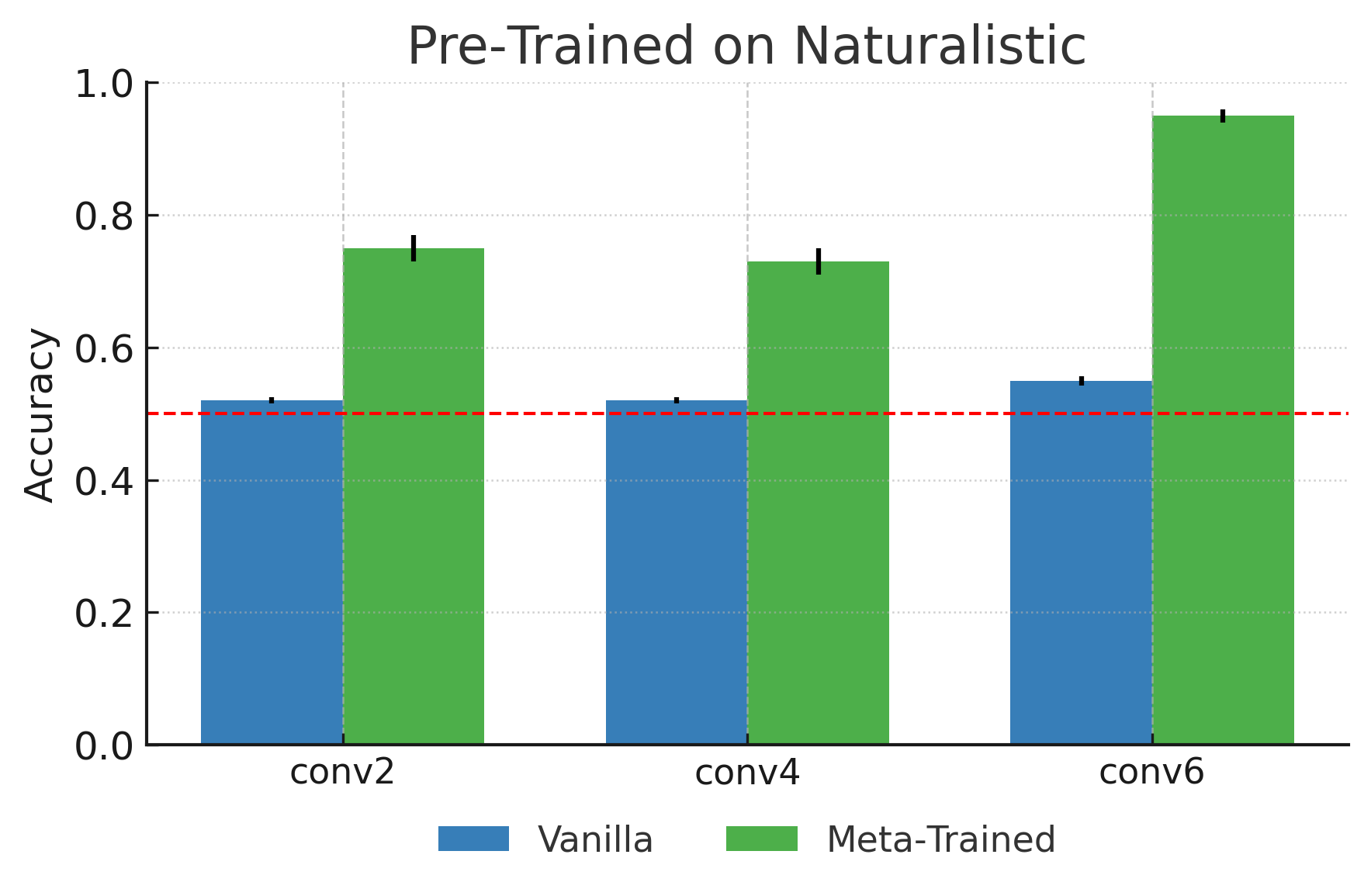}{2}{77}
    \label{fig:panelC}
  \end{subfigure}

  \caption{Generalizing the same-different relation to naturalistic stimuli. (A) Naturalistic examples of ``same’’ and ``different’’ images. (B) Model performance on naturalistic image data from Brady et al.~(2008), showing generalization performance for the models from the previous sections trained with MAML and conventional gradient descent on the data from Puebla and Bowers (2022) and tested on the 'naturalistic' data from Brady et al.~(2008). (C) Results of using the same model architectures and applying meta-learning or conventional learning on a subset of the naturalistic images and testing on unseen examples from the same dataset. An unseen example in this case represents a newly sampled object paired with either a translated or rotated version of that same object (a ``same'' task) or of a different, randomly selected object from the dataset (a ``different'' task). Error bars represent results averaged over five seeds.}
  \label{fig:brady_side_by_side}
\end{figure}

\begin{figure}[htb]
    \centering
    \includegraphics[width=0.8\textwidth]{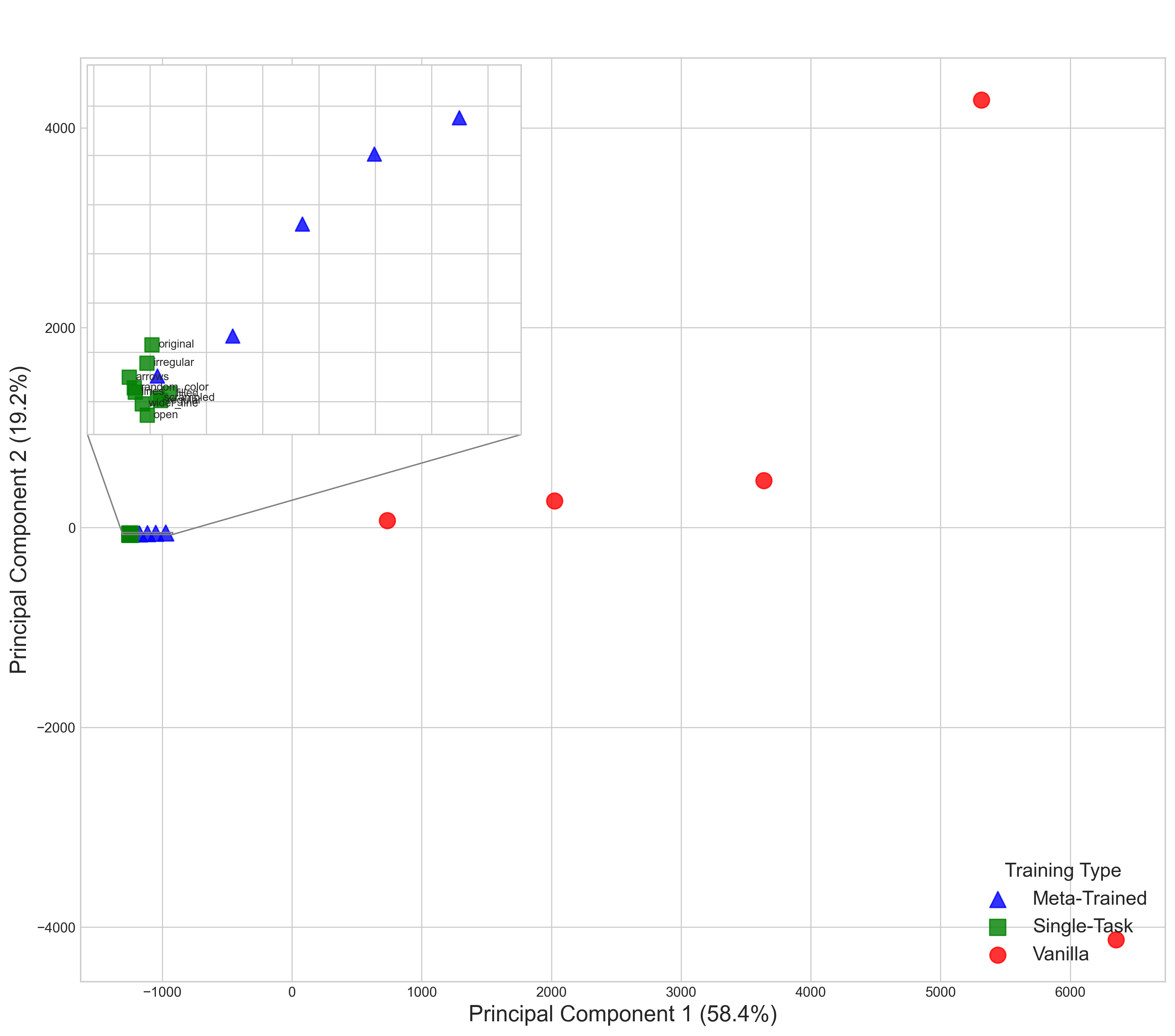}
    \caption{Meta-trained weights (blue triangles) lie closer in weight space to the optimal weights for each individual task (green squares each representing one task from Puebla and Bowers, 2022). The blue triangles are closer to the optimal weights than the random initial weights (red circles), helping to explain the rapid generalization abilities of the CNNs trained using MAML.}
    \label{fig:fig4-2}
\end{figure}

We compared the initial weights learned by MAML to the random initialization used in conventional learning (in both cases using five random seeds) and to the optimal weights for performing each task. The optimal weights were approximated by training a separate CNN network on each of the 10 same-different tasks from Puebla and Bowers with ten times the original amount of data, and allowing training until full convergence so that we ended up with a set of weights that were specifically fit to that task. We concatenated the weights of each neural network and flattened them into a single high-dimensional vector, giving us a point in high-dimensional space corresponding to each network. We then applied PCA to this set of vectors. 

The goal of MAML is to find a set of initial weights that minimizes the loss resulting from applying gradient descent from that initialization on each training task (see Appendix B for more details). Intuitively, this means that the initial weights learned by MAML should be close to the optimal weights for each.  Figure 7 shows that this is the case, with meta-learned initializations (blue triangles) being compared to random initializations (red circles) and the estimated optimal weights (green squares). We clearly see that meta-learned initializations are much closer in weight space to the per-task optima than the random weights. In other words, MAML has discovered a weight initialization from which a few gradient steps can reach the specialized solutions for every task, explaining the rapid generalization observed in our experiments. This provides strong evidence that the model has learned an effective inductive bias for solving the broad family of same-different tasks.


\section{Discussion}

We have shown that convolutional neural networks can learn to overcome longstanding limitations in relational reasoning if they are trained in the right way. By evaluating a simple class of networks on benchmarks of varying complexity, our results demonstrate that CNNs trained via meta-learning can learn a generalizable notion of the same-different relation. Through weight space analyses and systematic generalization tests, we provide evidence that these networks have internalized a version of the same-different relation that is abstract enough to generalize to new types of unseen objects.
While much work has recently focused on novel neural architectures dedicated to improving relational reasoning in vision models \citep{webb2021emergent,kerg2022inductive,altabaa2024abstractors,webb2024relational}, our results suggest that an alternative pathway to enhanced relational reasoning is via the nature of the training algorithm: optimizing a standard CNN (without architectural modifications) using meta-learning rather than standard learning. 

What is it about meta-learning that leads to the enhanced learning of the same-different relation we have observed? One possible explanation is that it may provide an incentive for abstraction. To understand this point, it is useful to compare meta-learning to conventional learning by stochastic gradient descent. In conventional learning, a network is shown a batch of examples, and its weights are then adjusted such that, if it were to process those same examples again, it would achieve a smaller error. 
In contrast, in meta-learning, each batch of examples (i.e., each episode) has two parts: the  labeled examples and the queries. A copy of the network is trained on the labeled examples and evaluated on the queries, and the original network's weights are then adjusted such that, if it were trained again on the labeled examples, its performance on the queries would improve.
Thus, meta-learning incentivizes the model to be able to learn from labeled examples in a way that is useful for processing new queries. 

Meta-learning creates a pressure for generalizing from one set of examples to another, which facilitates abstraction because low-level features (e.g., the angles of particular shapes) are unlikely to be broadly useful, whereas more abstract features (e.g., same-different information) will have more general utility. The MAML algorithm captures this ability to learn across examples through a second-order derivative, which, as we show in Appendix B, is causally implicated in model performance (models trained using first-order MAML, ignoring the second derivative, perform at chance level in the IID experiment). A derivation of first versus second order MAML is given in Appendix B. These results add to a growing body of evidence that meta-learning can help neural networks overcome some of their most troublesome limitations -- namely being able to learn from small amounts of data \citep{mccoy2023modeling} and demonstrating compositional generalization \citep{lake2023human}.


Our results support the idea that meta-learning allows models to overcome brittle memorization-based performance by allowing neural networks more opportunity to practice the type of behavior (relational reasoning) that we want to elicit. \citet{irie2024neural} call this the alignment of practice and incentive. By giving the networks the chance to be explicitly trained end-to-end on relational tasks of varying lengths and object types we create an incentive to generalize beyond the surface-level statistics of an individual training example. We hypothesize that this approach may also be useful for allowing neural networks to learn more robustly and efficiently in other tasks that demand complex relational and abstract reasoning. The same-different relation is key to the development of more complex relational reasoning. By showing the conditions for the emergence of this basic relation, our results suggest that meta-learning might enable neural networks to outperform conventional training methods in more challenging relational reasoning problems. 

Our training procedure also makes strong use of two ordering techniques from cognitive science that have proved useful for human learning: interleaving and blocking. Concatenating multiple labeled examples from a single task is an implicit form of the blocking technique: allowing inner loop adaptation on many similar examples with the same structure (see Appendix B for inner loop dynamics). On top of this implicit blocking structure that is natural to meta-learning, we also randomly intermix episodes of different task types during training, which provides the benefits of interleaving: when a rule is abstract or hard to learn independently, previous work has shown that interleaving examples from different but related tasks can improve recall and accuracy in humans \citep{Kornell2008, Carvalho2014, Goldstone1996}. Intriguingly, when we ablated the random interleaving effect (by presenting all episodes of a task together at the same time, followed by the next set of task-specific episodes sequentially), performance drops catastrophically. This similarity between human and neural network learning dynamics points to deeper underlying principles and deserves further investigation. 


\section{Conclusion}

Although a simple and intuitive relation, the same-different relation has proven notoriously difficult for neural networks to learn from data alone, suggesting that there may be some innate cognitive structures human learners bring to the problem that makes this a relatively trivial task. Questions surrounding the innateness of ``cognitive primitives'' have fueled much debate about what these structures may be and how we may replicate them in-silico \citep{spelke2007core}. For example, there is a debate about whether the notion of sameness is an innate cognitive primitive in humans and other animals \citep{gentner2021learning, premack1983codes}. In this work, we show that a careful restructuring of the training data itself into a meta-learning curriculum can obviate the need for in-built primitives or symbols: simply running gradient descent on appropriately structured training data is enough to generalize the same-different relation. This highlights an important trend in machine learning and cognitive science that stresses the importance of domain-general architectures that are able to meta-learn through domain-specific curricula \citep{bencomo2025teasing, mccoy2025modeling, lake2023human}. 

To a large extent, modern computer vision has advanced through optimizing 
classification objectives for individual objects, where objects are recognized by the statistical pattern generated by their individual pixels. 
What can be left out in the process is the rich ``invisible'' space between objects that humans seem to make sense of effortlessly: how individual objects relate to one another. Our work provides a path towards creating neural networks that are better able to learn relations across novel stimuli, using the higher-order nature of the gradient updates in meta-learning to encourage the development of more abstract relational information inside neural networks.


\newpage

\nocite{ChalnickBillman1988a}
\nocite{Feigenbaum1963a}
\nocite{Hill1983a}
\nocite{OhlssonLangley1985a}
\nocite{Matlock2001}
\nocite{NewellSimon1972a}
\nocite{ShragerLangley1990a}


\bibliographystyle{apacite}

\setlength{\bibleftmargin}{.125in}
\setlength{\bibindent}{-\bibleftmargin}


\begin{thebibliography}{}

\bibitem [\protect \citeauthoryear {%
Altabaa%
, Webb%
, Cohen%
\BCBL {}\ \BBA {} Lafferty%
}{%
Altabaa%
\ \protect \BOthers {.}}{%
{\protect \APACyear {2024}}%
}]{%
altabaa2024abstractors}
\APACinsertmetastar {%
altabaa2024abstractors}%
\begin{APACrefauthors}%
Altabaa, A.%
, Webb, T\BPBI W.%
, Cohen, J\BPBI D.%
\BCBL {}\ \BBA {} Lafferty, J.%
\end{APACrefauthors}%
\unskip\
\newblock
\APACrefYearMonthDay{2024}{}{}.
\newblock
{\BBOQ}\APACrefatitle {Abstractors and relational cross-attention: An inductive bias for explicit relational reasoning in Transformers} {Abstractors and relational cross-attention: An inductive bias for explicit relational reasoning in transformers}.{\BBCQ}
\newblock
\BIn{} \APACrefbtitle {{Proceedings of the Twelfth International Conference on Learning Representations}.} {{Proceedings of the Twelfth International Conference on Learning Representations}.}
\PrintBackRefs{\CurrentBib}

\bibitem [\protect \citeauthoryear {%
Alzubaidi%
\ \protect \BOthers {.}}{%
Alzubaidi%
\ \protect \BOthers {.}}{%
{\protect \APACyear {2021}}%
}]{%
alzubaidi2021review}
\APACinsertmetastar {%
alzubaidi2021review}%
\begin{APACrefauthors}%
Alzubaidi, L.%
, Zhang, J.%
, Humaidi, A\BPBI J.%
, Al-Dujaili, A.%
, Duan, Y.%
, Al-Shamma, O.%
\BDBL {}Farhan, L.%
\end{APACrefauthors}%
\unskip\
\newblock
\APACrefYearMonthDay{2021}{}{}.
\newblock
{\BBOQ}\APACrefatitle {Review of deep learning: concepts, {CNN} architectures, challenges, applications, future directions} {Review of deep learning: concepts, {CNN} architectures, challenges, applications, future directions}.{\BBCQ}
\newblock
\APACjournalVolNumPages{Journal of Big Data}{8}{}{1--74}.
\PrintBackRefs{\CurrentBib}

\bibitem [\protect \citeauthoryear {%
Antoniou%
, Edwards%
\BCBL {}\ \BBA {} Storkey%
}{%
Antoniou%
\ \protect \BOthers {.}}{%
{\protect \APACyear {2019}}%
}]{%
antoniou2018how}
\APACinsertmetastar {%
antoniou2018how}%
\begin{APACrefauthors}%
Antoniou, A.%
, Edwards, H.%
\BCBL {}\ \BBA {} Storkey, A.%
\end{APACrefauthors}%
\unskip\
\newblock
\APACrefYearMonthDay{2019}{}{}.
\newblock
{\BBOQ}\APACrefatitle {How to train your {MAML}} {How to train your {MAML}}.{\BBCQ}
\newblock
\BIn{} \APACrefbtitle {International Conference on Learning Representations.} {International conference on learning representations.}
\PrintBackRefs{\CurrentBib}

\bibitem [\protect \citeauthoryear {%
Bencomo%
, Gupta%
, Marinescu%
, McCoy%
\BCBL {}\ \BBA {} Griffiths%
}{%
Bencomo%
\ \protect \BOthers {.}}{%
{\protect \APACyear {2025}}%
}]{%
bencomo2025teasing}
\APACinsertmetastar {%
bencomo2025teasing}%
\begin{APACrefauthors}%
Bencomo, G.%
, Gupta, M.%
, Marinescu, I.%
, McCoy, R\BPBI T.%
\BCBL {}\ \BBA {} Griffiths, T\BPBI L.%
\end{APACrefauthors}%
\unskip\
\newblock
\APACrefYearMonthDay{2025}{{\APACmonth{02}}}{}.
\newblock
\APACrefbtitle {Teasing Apart Architecture and Initial Weights as Sources of Inductive Bias in Neural Networks.} {Teasing apart architecture and initial weights as sources of inductive bias in neural networks.}
\newblock
\APAChowpublished {\url{https://arxiv.org/abs/2502.20237}}.
\newblock
\APACrefnote{arXiv preprint arXiv:2502.20237}
\PrintBackRefs{\CurrentBib}

\bibitem [\protect \citeauthoryear {%
Bl{\"o}te%
, Resing%
, Mazer%
\BCBL {}\ \BBA {} Van~Noort%
}{%
Bl{\"o}te%
\ \protect \BOthers {.}}{%
{\protect \APACyear {1999}}%
}]{%
blote1999young}
\APACinsertmetastar {%
blote1999young}%
\begin{APACrefauthors}%
Bl{\"o}te, A\BPBI W.%
, Resing, W\BPBI C.%
, Mazer, P.%
\BCBL {}\ \BBA {} Van~Noort, D\BPBI A.%
\end{APACrefauthors}%
\unskip\
\newblock
\APACrefYearMonthDay{1999}{}{}.
\newblock
{\BBOQ}\APACrefatitle {Young children's organizational strategies on a same--different task: A microgenetic study and a training study} {Young children's organizational strategies on a same--different task: A microgenetic study and a training study}.{\BBCQ}
\newblock
\APACjournalVolNumPages{Journal of Experimental Child Psychology}{74}{1}{21--43}.
\PrintBackRefs{\CurrentBib}

\bibitem [\protect \citeauthoryear {%
Brady%
, Konkle%
, Alvarez%
\BCBL {}\ \BBA {} Oliva%
}{%
Brady%
\ \protect \BOthers {.}}{%
{\protect \APACyear {2008}}%
}]{%
brady2008visual}
\APACinsertmetastar {%
brady2008visual}%
\begin{APACrefauthors}%
Brady, T\BPBI F.%
, Konkle, T.%
, Alvarez, G\BPBI A.%
\BCBL {}\ \BBA {} Oliva, A.%
\end{APACrefauthors}%
\unskip\
\newblock
\APACrefYearMonthDay{2008}{{\APACmonth{09}}}{}.
\newblock
{\BBOQ}\APACrefatitle {Visual long-term memory has a massive storage capacity for object details} {Visual long-term memory has a massive storage capacity for object details}.{\BBCQ}
\newblock
\APACjournalVolNumPages{Proceedings of the National Academy of Sciences of the United States of America}{105}{38}{14325--14329}.
\PrintBackRefs{\CurrentBib}

\bibitem [\protect \citeauthoryear {%
Carvalho%
\ \BBA {} Goldstone%
}{%
Carvalho%
\ \BBA {} Goldstone%
}{%
{\protect \APACyear {2014}}%
}]{%
Carvalho2014}
\APACinsertmetastar {%
Carvalho2014}%
\begin{APACrefauthors}%
Carvalho, P\BPBI F.%
\BCBT {}\ \BBA {} Goldstone, R\BPBI L.%
\end{APACrefauthors}%
\unskip\
\newblock
\APACrefYearMonthDay{2014}{}{}.
\newblock
{\BBOQ}\APACrefatitle {What you learn is more than what you see: What can sequencing effects tell us about inductive category learning?} {What you learn is more than what you see: What can sequencing effects tell us about inductive category learning?}{\BBCQ}
\newblock
\APACjournalVolNumPages{Cognitive Psychology}{69}{}{1--25}.
\PrintBackRefs{\CurrentBib}

\bibitem [\protect \citeauthoryear {%
Chalnick%
\ \BBA {} Billman%
}{%
Chalnick%
\ \BBA {} Billman%
}{%
{\protect \APACyear {1988}}%
}]{%
ChalnickBillman1988a}
\APACinsertmetastar {%
ChalnickBillman1988a}%
\begin{APACrefauthors}%
Chalnick, A.%
\BCBT {}\ \BBA {} Billman, D.%
\end{APACrefauthors}%
\unskip\
\newblock
\APACrefYearMonthDay{1988}{}{}.
\newblock
{\BBOQ}\APACrefatitle {Unsupervised learning of correlational structure} {Unsupervised learning of correlational structure}.{\BBCQ}
\newblock
\BIn{} \APACrefbtitle {{Proceedings of the Tenth Annual Conference of the Cognitive Science Society}} {{Proceedings of the Tenth Annual Conference of the Cognitive Science Society}}\ (\BPGS\ 510--516).
\newblock
\APACaddressPublisher{Hillsdale, NJ}{Lawrence Erlbaum Associates}.
\PrintBackRefs{\CurrentBib}

\bibitem [\protect \citeauthoryear {%
Dosovitskiy%
}{%
Dosovitskiy%
}{%
{\protect \APACyear {2020}}%
}]{%
dosovitskiy2020image}
\APACinsertmetastar {%
dosovitskiy2020image}%
\begin{APACrefauthors}%
Dosovitskiy, A.%
\end{APACrefauthors}%
\unskip\
\newblock
\APACrefYearMonthDay{2020}{}{}.
\newblock
{\BBOQ}\APACrefatitle {An image is worth $16 \times 16$ words: Transformers for image recognition at scale} {An image is worth $16 \times 16$ words: Transformers for image recognition at scale}.{\BBCQ}
\newblock
\APACjournalVolNumPages{arXiv preprint arXiv:2010.11929}{}{}{}.
\PrintBackRefs{\CurrentBib}

\bibitem [\protect \citeauthoryear {%
Feigenbaum%
}{%
Feigenbaum%
}{%
{\protect \APACyear {1963}}%
}]{%
Feigenbaum1963a}
\APACinsertmetastar {%
Feigenbaum1963a}%
\begin{APACrefauthors}%
Feigenbaum, E\BPBI A.%
\end{APACrefauthors}%
\unskip\
\newblock
\APACrefYearMonthDay{1963}{}{}.
\newblock
{\BBOQ}\APACrefatitle {The simulation of verbal learning behavior} {The simulation of verbal learning behavior}.{\BBCQ}
\newblock
\BIn{} E\BPBI A.~Feigenbaum\ \BBA {} J.~Feldman\ (\BEDS), \APACrefbtitle {Computers and Thought.} {Computers and thought.}
\newblock
\APACaddressPublisher{New York}{McGraw-Hill}.
\PrintBackRefs{\CurrentBib}

\bibitem [\protect \citeauthoryear {%
Finn%
, Abbeel%
\BCBL {}\ \BBA {} Levine%
}{%
Finn%
\ \protect \BOthers {.}}{%
{\protect \APACyear {2017}}%
}]{%
finn2017model}
\APACinsertmetastar {%
finn2017model}%
\begin{APACrefauthors}%
Finn, C.%
, Abbeel, P.%
\BCBL {}\ \BBA {} Levine, S.%
\end{APACrefauthors}%
\unskip\
\newblock
\APACrefYearMonthDay{2017}{}{}.
\newblock
{\BBOQ}\APACrefatitle {Model-agnostic meta-learning for fast adaptation of deep networks} {Model-agnostic meta-learning for fast adaptation of deep networks}.{\BBCQ}
\newblock
\BIn{} \APACrefbtitle {{International Conference on Machine Learning}} {{International Conference on Machine Learning}}\ (\BPGS\ 1126--1135).
\PrintBackRefs{\CurrentBib}

\bibitem [\protect \citeauthoryear {%
Fleuret%
\ \protect \BOthers {.}}{%
Fleuret%
\ \protect \BOthers {.}}{%
{\protect \APACyear {2011}}%
}]{%
fleuret2011comparing}
\APACinsertmetastar {%
fleuret2011comparing}%
\begin{APACrefauthors}%
Fleuret, F.%
, Li, T.%
, Dubout, C.%
, Wampler, E\BPBI K.%
, Yantis, S.%
\BCBL {}\ \BBA {} Geman, D.%
\end{APACrefauthors}%
\unskip\
\newblock
\APACrefYearMonthDay{2011}{}{}.
\newblock
{\BBOQ}\APACrefatitle {Comparing machines and humans on a visual categorization test} {Comparing machines and humans on a visual categorization test}.{\BBCQ}
\newblock
\APACjournalVolNumPages{Proceedings of the National Academy of Sciences}{108}{43}{17621--17625}.
\PrintBackRefs{\CurrentBib}

\bibitem [\protect \citeauthoryear {%
Fodor%
\ \BBA {} Pylyshyn%
}{%
Fodor%
\ \BBA {} Pylyshyn%
}{%
{\protect \APACyear {1988}}%
}]{%
fodor1988connectionism}
\APACinsertmetastar {%
fodor1988connectionism}%
\begin{APACrefauthors}%
Fodor, J\BPBI A.%
\BCBT {}\ \BBA {} Pylyshyn, Z\BPBI W.%
\end{APACrefauthors}%
\unskip\
\newblock
\APACrefYearMonthDay{1988}{}{}.
\newblock
{\BBOQ}\APACrefatitle {Connectionism and cognitive architecture: A critical analysis} {Connectionism and cognitive architecture: A critical analysis}.{\BBCQ}
\newblock
\APACjournalVolNumPages{Cognition}{28}{1-2}{3--71}.
\PrintBackRefs{\CurrentBib}

\bibitem [\protect \citeauthoryear {%
Gentner%
, Shao%
, Simms%
\BCBL {}\ \BBA {} Hespos%
}{%
Gentner%
\ \protect \BOthers {.}}{%
{\protect \APACyear {2021}}%
}]{%
gentner2021learning}
\APACinsertmetastar {%
gentner2021learning}%
\begin{APACrefauthors}%
Gentner, D.%
, Shao, R.%
, Simms, N.%
\BCBL {}\ \BBA {} Hespos, S.%
\end{APACrefauthors}%
\unskip\
\newblock
\APACrefYearMonthDay{2021}{}{}.
\newblock
{\BBOQ}\APACrefatitle {Learning same and different relations: cross-species comparisons} {Learning same and different relations: cross-species comparisons}.{\BBCQ}
\newblock
\APACjournalVolNumPages{Current Opinion in Behavioral Sciences}{37}{}{84--89}.
\PrintBackRefs{\CurrentBib}

\bibitem [\protect \citeauthoryear {%
Goldstone%
}{%
Goldstone%
}{%
{\protect \APACyear {1996}}%
}]{%
Goldstone1996}
\APACinsertmetastar {%
Goldstone1996}%
\begin{APACrefauthors}%
Goldstone, R\BPBI L.%
\end{APACrefauthors}%
\unskip\
\newblock
\APACrefYearMonthDay{1996}{}{}.
\newblock
{\BBOQ}\APACrefatitle {Isolated and Interacting Category Representations} {Isolated and interacting category representations}.{\BBCQ}
\newblock
\BIn{} B\BPBI H.~Ross\ (\BED), \APACrefbtitle {Psychology of Learning and Motivation} {Psychology of learning and motivation}\ (\BVOL~36, \BPGS\ 65--116).
\newblock
\APACaddressPublisher{}{Academic Press}.
\PrintBackRefs{\CurrentBib}

\bibitem [\protect \citeauthoryear {%
Gupta%
, Rane%
, McCoy%
\BCBL {}\ \BBA {} Griffiths%
}{%
Gupta%
\ \protect \BOthers {.}}{%
{\protect \APACyear {2025}}%
}]{%
gupta2025convolutional}
\APACinsertmetastar {%
gupta2025convolutional}%
\begin{APACrefauthors}%
Gupta, M.%
, Rane, S.%
, McCoy, R\BPBI T.%
\BCBL {}\ \BBA {} Griffiths, T\BPBI L.%
\end{APACrefauthors}%
\unskip\
\newblock
\APACrefYearMonthDay{2025}{}{}.
\newblock
{\BBOQ}\APACrefatitle {Convolutional Neural Networks Can (Meta-) Learn the Same-Different Relation} {Convolutional neural networks can (meta-) learn the same-different relation}.{\BBCQ}
\newblock
\BIn{} \APACrefbtitle {{Proceedings of the 44th Annual Conference of the Cognitive Science Society}.} {{Proceedings of the 44th Annual Conference of the Cognitive Science Society}.}
\PrintBackRefs{\CurrentBib}

\bibitem [\protect \citeauthoryear {%
Hasson%
, Nastase%
\BCBL {}\ \BBA {} Goldstein%
}{%
Hasson%
\ \protect \BOthers {.}}{%
{\protect \APACyear {2020}}%
}]{%
hasson2020direct}
\APACinsertmetastar {%
hasson2020direct}%
\begin{APACrefauthors}%
Hasson, U.%
, Nastase, S\BPBI A.%
\BCBL {}\ \BBA {} Goldstein, A.%
\end{APACrefauthors}%
\unskip\
\newblock
\APACrefYearMonthDay{2020}{{\APACmonth{02}}}{}.
\newblock
{\BBOQ}\APACrefatitle {Direct Fit to Nature: An Evolutionary Perspective on Biological and Artificial Neural Networks} {Direct fit to nature: An evolutionary perspective on biological and artificial neural networks}.{\BBCQ}
\newblock
\APACjournalVolNumPages{Neuron}{105}{3}{416--434}.
\newblock
\begin{APACrefDOI} \doi{10.1016/j.neuron.2019.12.002} \end{APACrefDOI}
\PrintBackRefs{\CurrentBib}

\bibitem [\protect \citeauthoryear {%
He%
, Zhang%
, Ren%
\BCBL {}\ \BBA {} Sun%
}{%
He%
\ \protect \BOthers {.}}{%
{\protect \APACyear {2016}}%
}]{%
he2016deep}
\APACinsertmetastar {%
he2016deep}%
\begin{APACrefauthors}%
He, K.%
, Zhang, X.%
, Ren, S.%
\BCBL {}\ \BBA {} Sun, J.%
\end{APACrefauthors}%
\unskip\
\newblock
\APACrefYearMonthDay{2016}{}{}.
\newblock
{\BBOQ}\APACrefatitle {Deep residual learning for image recognition} {Deep residual learning for image recognition}.{\BBCQ}
\newblock
\BIn{} \APACrefbtitle {Proceedings of the {IEEE} {C}onference on {C}omputer {V}ision and {P}attern {R}ecognition {(CVPR)}} {Proceedings of the {IEEE} {C}onference on {C}omputer {V}ision and {P}attern {R}ecognition {(CVPR)}}\ (\BPGS\ 770--778).
\PrintBackRefs{\CurrentBib}

\bibitem [\protect \citeauthoryear {%
Hill%
}{%
Hill%
}{%
{\protect \APACyear {1983}}%
}]{%
Hill1983a}
\APACinsertmetastar {%
Hill1983a}%
\begin{APACrefauthors}%
Hill, J\BPBI A\BPBI C.%
\end{APACrefauthors}%
\unskip\
\newblock
\APACrefYearMonthDay{1983}{}{}.
\newblock
{\BBOQ}\APACrefatitle {A computational model of language acquisition in the two-year old} {A computational model of language acquisition in the two-year old}.{\BBCQ}
\newblock
\APACjournalVolNumPages{Cognition and Brain Theory}{6}{}{287--317}.
\PrintBackRefs{\CurrentBib}

\bibitem [\protect \citeauthoryear {%
Hubel%
, Wiesel%
\BCBL {}\ \protect \BOthers {.}}{%
Hubel%
\ \protect \BOthers {.}}{%
{\protect \APACyear {1959}}%
}]{%
hubel1959receptive}
\APACinsertmetastar {%
hubel1959receptive}%
\begin{APACrefauthors}%
Hubel, D\BPBI H.%
, Wiesel, T\BPBI N.%
\BCBL {}\ \BOthersPeriod {.}\end{APACrefauthors}%
\unskip\
\newblock
\APACrefYearMonthDay{1959}{}{}.
\newblock
{\BBOQ}\APACrefatitle {Receptive fields of single neurones in the cat’s striate cortex} {Receptive fields of single neurones in the cat’s striate cortex}.{\BBCQ}
\newblock
\APACjournalVolNumPages{Journal of Physiology}{148}{3}{574--591}.
\PrintBackRefs{\CurrentBib}

\bibitem [\protect \citeauthoryear {%
Irie%
\ \BBA {} Lake%
}{%
Irie%
\ \BBA {} Lake%
}{%
{\protect \APACyear {2024}}%
}]{%
irie2024neural}
\APACinsertmetastar {%
irie2024neural}%
\begin{APACrefauthors}%
Irie, K.%
\BCBT {}\ \BBA {} Lake, B\BPBI M.%
\end{APACrefauthors}%
\unskip\
\newblock
\APACrefYearMonthDay{2024}{}{}.
\newblock
{\BBOQ}\APACrefatitle {Neural networks that overcome classic challenges through practice} {Neural networks that overcome classic challenges through practice}.{\BBCQ}
\newblock
\APACjournalVolNumPages{arXiv preprint arXiv:2410.10596}{}{}{}.
\PrintBackRefs{\CurrentBib}

\bibitem [\protect \citeauthoryear {%
Kerg%
\ \protect \BOthers {.}}{%
Kerg%
\ \protect \BOthers {.}}{%
{\protect \APACyear {2022}}%
}]{%
kerg2022inductive}
\APACinsertmetastar {%
kerg2022inductive}%
\begin{APACrefauthors}%
Kerg, G.%
, Mittal, S.%
, Rolnick, D.%
, Bengio, Y.%
, Richards, B\BPBI A.%
\BCBL {}\ \BBA {} Lajoie, G.%
\end{APACrefauthors}%
\unskip\
\newblock
\APACrefYearMonthDay{2022}{}{}.
\newblock
{\BBOQ}\APACrefatitle {Inductive Biases for Relational Tasks} {Inductive biases for relational tasks}.{\BBCQ}
\newblock
\BIn{} \APACrefbtitle {{ICLR 2022 Workshop on the Elements of Reasoning: Objects, Structure and Causality}.} {{ICLR 2022 Workshop on the Elements of Reasoning: Objects, Structure and Causality}.}
\PrintBackRefs{\CurrentBib}

\bibitem [\protect \citeauthoryear {%
Kim%
, Ricci%
\BCBL {}\ \BBA {} Serre%
}{%
Kim%
\ \protect \BOthers {.}}{%
{\protect \APACyear {2018}}%
}]{%
kim2018not}
\APACinsertmetastar {%
kim2018not}%
\begin{APACrefauthors}%
Kim, J.%
, Ricci, M.%
\BCBL {}\ \BBA {} Serre, T.%
\end{APACrefauthors}%
\unskip\
\newblock
\APACrefYearMonthDay{2018}{}{}.
\newblock
{\BBOQ}\APACrefatitle {Not-So-{CLEVR}: learning same--different relations strains feedforward neural networks} {Not-so-{CLEVR}: learning same--different relations strains feedforward neural networks}.{\BBCQ}
\newblock
\APACjournalVolNumPages{Interface Focus}{8}{4}{20180011}.
\PrintBackRefs{\CurrentBib}

\bibitem [\protect \citeauthoryear {%
Kingma%
\ \BBA {} Ba%
}{%
Kingma%
\ \BBA {} Ba%
}{%
{\protect \APACyear {2015}}%
}]{%
kingma2015adam}
\APACinsertmetastar {%
kingma2015adam}%
\begin{APACrefauthors}%
Kingma, D.%
\BCBT {}\ \BBA {} Ba, J.%
\end{APACrefauthors}%
\unskip\
\newblock
\APACrefYearMonthDay{2015}{}{}.
\newblock
{\BBOQ}\APACrefatitle {Adam: A Method for Stochastic Optimization} {Adam: A method for stochastic optimization}.{\BBCQ}
\newblock
\BIn{} \APACrefbtitle {International {C}onference on {L}earning {R}epresentations.} {International {C}onference on {L}earning {R}epresentations.}
\PrintBackRefs{\CurrentBib}

\bibitem [\protect \citeauthoryear {%
Kornell%
\ \BBA {} Bjork%
}{%
Kornell%
\ \BBA {} Bjork%
}{%
{\protect \APACyear {2008}}%
}]{%
Kornell2008}
\APACinsertmetastar {%
Kornell2008}%
\begin{APACrefauthors}%
Kornell, N.%
\BCBT {}\ \BBA {} Bjork, R\BPBI A.%
\end{APACrefauthors}%
\unskip\
\newblock
\APACrefYearMonthDay{2008}{}{}.
\newblock
{\BBOQ}\APACrefatitle {Learning Concepts and Categories: Is Spacing the "Enemy of Induction"?} {Learning concepts and categories: Is spacing the "enemy of induction"?}{\BBCQ}
\newblock
\APACjournalVolNumPages{Psychological Science}{19}{6}{585--592}.
\newblock
\begin{APACrefDOI} \doi{10.1111/j.1467-9280.2008.02127.x} \end{APACrefDOI}
\PrintBackRefs{\CurrentBib}

\bibitem [\protect \citeauthoryear {%
Krizhevsky%
, Sutskever%
\BCBL {}\ \BBA {} Hinton%
}{%
Krizhevsky%
\ \protect \BOthers {.}}{%
{\protect \APACyear {2012}}%
}]{%
krizhevsky2012imagenet}
\APACinsertmetastar {%
krizhevsky2012imagenet}%
\begin{APACrefauthors}%
Krizhevsky, A.%
, Sutskever, I.%
\BCBL {}\ \BBA {} Hinton, G\BPBI E.%
\end{APACrefauthors}%
\unskip\
\newblock
\APACrefYearMonthDay{2012}{}{}.
\newblock
{\BBOQ}\APACrefatitle {{ImageNet} classification with deep convolutional neural networks} {{ImageNet} classification with deep convolutional neural networks}.{\BBCQ}
\newblock
\APACjournalVolNumPages{Advances in {N}eural {I}nformation {P}rocessing {S}ystems}{25}{}{}.
\PrintBackRefs{\CurrentBib}

\bibitem [\protect \citeauthoryear {%
Kubilius%
\ \protect \BOthers {.}}{%
Kubilius%
\ \protect \BOthers {.}}{%
{\protect \APACyear {2019}}%
}]{%
kubilius2019brain}
\APACinsertmetastar {%
kubilius2019brain}%
\begin{APACrefauthors}%
Kubilius, J.%
, Schrimpf, M.%
, Kar, K.%
, Rajalingham, R.%
, Hong, H.%
, Majaj, N.%
\BDBL {}others%
\end{APACrefauthors}%
\unskip\
\newblock
\APACrefYearMonthDay{2019}{}{}.
\newblock
{\BBOQ}\APACrefatitle {Brain-like object recognition with high-performing shallow recurrent ANNs} {Brain-like object recognition with high-performing shallow recurrent anns}.{\BBCQ}
\newblock
\APACjournalVolNumPages{Advances in {N}eural {I}nformation {P}rocessing {S}ystems}{32}{}{}.
\PrintBackRefs{\CurrentBib}

\bibitem [\protect \citeauthoryear {%
Lake%
\ \BBA {} Baroni%
}{%
Lake%
\ \BBA {} Baroni%
}{%
{\protect \APACyear {2023}}%
}]{%
lake2023human}
\APACinsertmetastar {%
lake2023human}%
\begin{APACrefauthors}%
Lake, B\BPBI M.%
\BCBT {}\ \BBA {} Baroni, M.%
\end{APACrefauthors}%
\unskip\
\newblock
\APACrefYearMonthDay{2023}{}{}.
\newblock
{\BBOQ}\APACrefatitle {Human-like systematic generalization through a meta-learning neural network} {Human-like systematic generalization through a meta-learning neural network}.{\BBCQ}
\newblock
\APACjournalVolNumPages{Nature}{623}{7985}{115--121}.
\PrintBackRefs{\CurrentBib}

\bibitem [\protect \citeauthoryear {%
LeCun%
, Bengio%
\BCBL {}\ \BBA {} Hinton%
}{%
LeCun%
\ \protect \BOthers {.}}{%
{\protect \APACyear {2015}}%
}]{%
lecun2015deep}
\APACinsertmetastar {%
lecun2015deep}%
\begin{APACrefauthors}%
LeCun, Y.%
, Bengio, Y.%
\BCBL {}\ \BBA {} Hinton, G.%
\end{APACrefauthors}%
\unskip\
\newblock
\APACrefYearMonthDay{2015}{}{}.
\newblock
{\BBOQ}\APACrefatitle {Deep learning} {Deep learning}.{\BBCQ}
\newblock
\APACjournalVolNumPages{Nature}{521}{7553}{436--444}.
\PrintBackRefs{\CurrentBib}

\bibitem [\protect \citeauthoryear {%
Lupker%
, Nakayama%
\BCBL {}\ \BBA {} Perea%
}{%
Lupker%
\ \protect \BOthers {.}}{%
{\protect \APACyear {2015}}%
}]{%
lupker2015there}
\APACinsertmetastar {%
lupker2015there}%
\begin{APACrefauthors}%
Lupker, S\BPBI J.%
, Nakayama, M.%
\BCBL {}\ \BBA {} Perea, M.%
\end{APACrefauthors}%
\unskip\
\newblock
\APACrefYearMonthDay{2015}{}{}.
\newblock
{\BBOQ}\APACrefatitle {Is there phonologically based priming in the same-different task? {E}vidence from {Japanese-English} bilinguals.} {Is there phonologically based priming in the same-different task? {E}vidence from {Japanese-English} bilinguals.}{\BBCQ}
\newblock
\APACjournalVolNumPages{Journal of Experimental Psychology: Human Perception and Performance}{41}{5}{1281}.
\PrintBackRefs{\CurrentBib}

\bibitem [\protect \citeauthoryear {%
Marinescu%
, McCoy%
\BCBL {}\ \BBA {} Griffiths%
}{%
Marinescu%
\ \protect \BOthers {.}}{%
{\protect \APACyear {2024}}%
}]{%
marinescu2024distilling}
\APACinsertmetastar {%
marinescu2024distilling}%
\begin{APACrefauthors}%
Marinescu, I.%
, McCoy, R\BPBI T.%
\BCBL {}\ \BBA {} Griffiths, T\BPBI L.%
\end{APACrefauthors}%
\unskip\
\newblock
\APACrefYearMonthDay{2024}{}{}.
\newblock
{\BBOQ}\APACrefatitle {Distilling symbolic priors for concept learning into neural networks} {Distilling symbolic priors for concept learning into neural networks}.{\BBCQ}
\newblock
\BIn{} \APACrefbtitle {{Proceedings of the 46th Annual Conference of the Cognitive Science Society}.} {{Proceedings of the 46th Annual Conference of the Cognitive Science Society}.}
\PrintBackRefs{\CurrentBib}

\bibitem [\protect \citeauthoryear {%
Matlock%
}{%
Matlock%
}{%
{\protect \APACyear {2001}}%
}]{%
Matlock2001}
\APACinsertmetastar {%
Matlock2001}%
\begin{APACrefauthors}%
Matlock, T.%
\end{APACrefauthors}%
\unskip\
\newblock
\APACrefYear{2001}.
\newblock
\APACrefbtitle {How real is fictive motion?} {How real is fictive motion?}
\newblock
Doctoral dissertation, Psychology Department, University of California, Santa Cruz.
\PrintBackRefs{\CurrentBib}

\bibitem [\protect \citeauthoryear {%
McCoy%
\ \BBA {} Griffiths%
}{%
McCoy%
\ \BBA {} Griffiths%
}{%
{\protect \APACyear {2023}}%
}]{%
mccoy2023modeling}
\APACinsertmetastar {%
mccoy2023modeling}%
\begin{APACrefauthors}%
McCoy, R\BPBI T.%
\BCBT {}\ \BBA {} Griffiths, T\BPBI L.%
\end{APACrefauthors}%
\unskip\
\newblock
\APACrefYearMonthDay{2023}{}{}.
\newblock
{\BBOQ}\APACrefatitle {Modeling rapid language learning by distilling {B}ayesian priors into artificial neural networks} {Modeling rapid language learning by distilling {B}ayesian priors into artificial neural networks}.{\BBCQ}
\newblock
\APACjournalVolNumPages{arXiv preprint arXiv:2305.14701}{}{}{}.
\PrintBackRefs{\CurrentBib}

\bibitem [\protect \citeauthoryear {%
McCoy%
\ \BBA {} Griffiths%
}{%
McCoy%
\ \BBA {} Griffiths%
}{%
{\protect \APACyear {2025}}%
}]{%
mccoy2025modeling}
\APACinsertmetastar {%
mccoy2025modeling}%
\begin{APACrefauthors}%
McCoy, R\BPBI T.%
\BCBT {}\ \BBA {} Griffiths, T\BPBI L.%
\end{APACrefauthors}%
\unskip\
\newblock
\APACrefYearMonthDay{2025}{}{}.
\newblock
{\BBOQ}\APACrefatitle {Modeling rapid language learning by distilling {B}ayesian priors into artificial neural networks} {Modeling rapid language learning by distilling {B}ayesian priors into artificial neural networks}.{\BBCQ}
\newblock
\APACjournalVolNumPages{Nature Communications}{16}{1}{4676}.
\PrintBackRefs{\CurrentBib}

\bibitem [\protect \citeauthoryear {%
Minsky%
\ \BBA {} Papert%
}{%
Minsky%
\ \BBA {} Papert%
}{%
{\protect \APACyear {1969}}%
}]{%
minsky1969introduction}
\APACinsertmetastar {%
minsky1969introduction}%
\begin{APACrefauthors}%
Minsky, M\BPBI L.%
\BCBT {}\ \BBA {} Papert, S\BPBI A.%
\end{APACrefauthors}%
\unskip\
\newblock
\APACrefYear{1969}.
\newblock
\APACrefbtitle {Perceptrons: An introduction to computational geometry} {Perceptrons: An introduction to computational geometry}.
\newblock
\APACaddressPublisher{Cambridge, MA}{MIT Press}.
\PrintBackRefs{\CurrentBib}

\bibitem [\protect \citeauthoryear {%
Newell%
\ \BBA {} Simon%
}{%
Newell%
\ \BBA {} Simon%
}{%
{\protect \APACyear {1972}}%
}]{%
NewellSimon1972a}
\APACinsertmetastar {%
NewellSimon1972a}%
\begin{APACrefauthors}%
Newell, A.%
\BCBT {}\ \BBA {} Simon, H\BPBI A.%
\end{APACrefauthors}%
\unskip\
\newblock
\APACrefYear{1972}.
\newblock
\APACrefbtitle {Human problem solving} {Human problem solving}.
\newblock
\APACaddressPublisher{Englewood Cliffs, NJ}{Prentice-Hall}.
\PrintBackRefs{\CurrentBib}

\bibitem [\protect \citeauthoryear {%
Ohlsson%
\ \BBA {} Langley%
}{%
Ohlsson%
\ \BBA {} Langley%
}{%
{\protect \APACyear {1985}}%
}]{%
OhlssonLangley1985a}
\APACinsertmetastar {%
OhlssonLangley1985a}%
\begin{APACrefauthors}%
Ohlsson, S.%
\BCBT {}\ \BBA {} Langley, P.%
\end{APACrefauthors}%
\unskip\
\newblock
\APACrefYearMonthDay{1985}{}{}.
\newblock
\APACrefbtitle {Identifying solution paths in cognitive diagnosis} {Identifying solution paths in cognitive diagnosis}\ \APACbVolEdTR{}{\BTR{}\ \BNUM\ CMU-RI-TR-85-2}.
\newblock
\APACaddressInstitution{Pittsburgh, PA}{Carnegie Mellon University, The Robotics Institute}.
\PrintBackRefs{\CurrentBib}

\bibitem [\protect \citeauthoryear {%
Peterson%
, Abbott%
\BCBL {}\ \BBA {} Griffiths%
}{%
Peterson%
\ \protect \BOthers {.}}{%
{\protect \APACyear {2018}}%
}]{%
peterson2018evaluating}
\APACinsertmetastar {%
peterson2018evaluating}%
\begin{APACrefauthors}%
Peterson, J\BPBI C.%
, Abbott, J\BPBI T.%
\BCBL {}\ \BBA {} Griffiths, T\BPBI L.%
\end{APACrefauthors}%
\unskip\
\newblock
\APACrefYearMonthDay{2018}{}{}.
\newblock
{\BBOQ}\APACrefatitle {Evaluating (and improving) the correspondence between deep neural networks and human representations} {Evaluating (and improving) the correspondence between deep neural networks and human representations}.{\BBCQ}
\newblock
\APACjournalVolNumPages{Cognitive Science}{42}{8}{2648--2669}.
\PrintBackRefs{\CurrentBib}

\bibitem [\protect \citeauthoryear {%
Pinker%
\ \BBA {} Prince%
}{%
Pinker%
\ \BBA {} Prince%
}{%
{\protect \APACyear {1988}}%
}]{%
pinker1988language}
\APACinsertmetastar {%
pinker1988language}%
\begin{APACrefauthors}%
Pinker, S.%
\BCBT {}\ \BBA {} Prince, A.%
\end{APACrefauthors}%
\unskip\
\newblock
\APACrefYearMonthDay{1988}{}{}.
\newblock
{\BBOQ}\APACrefatitle {On language and connectionism: Analysis of a parallel distributed processing model of language acquisition} {On language and connectionism: Analysis of a parallel distributed processing model of language acquisition}.{\BBCQ}
\newblock
\APACjournalVolNumPages{Cognition}{28}{1-2}{73--193}.
\PrintBackRefs{\CurrentBib}

\bibitem [\protect \citeauthoryear {%
Premack%
}{%
Premack%
}{%
{\protect \APACyear {1983}}%
}]{%
premack1983codes}
\APACinsertmetastar {%
premack1983codes}%
\begin{APACrefauthors}%
Premack, D.%
\end{APACrefauthors}%
\unskip\
\newblock
\APACrefYearMonthDay{1983}{}{}.
\newblock
{\BBOQ}\APACrefatitle {The codes of man and beasts} {The codes of man and beasts}.{\BBCQ}
\newblock
\APACjournalVolNumPages{Behavioral and Brain Sciences}{6}{1}{125--136}.
\PrintBackRefs{\CurrentBib}

\bibitem [\protect \citeauthoryear {%
Puebla%
\ \BBA {} Bowers%
}{%
Puebla%
\ \BBA {} Bowers%
}{%
{\protect \APACyear {2022}}%
}]{%
puebla2022can}
\APACinsertmetastar {%
puebla2022can}%
\begin{APACrefauthors}%
Puebla, G.%
\BCBT {}\ \BBA {} Bowers, J\BPBI S.%
\end{APACrefauthors}%
\unskip\
\newblock
\APACrefYearMonthDay{2022}{}{}.
\newblock
{\BBOQ}\APACrefatitle {Can deep convolutional neural networks support relational reasoning in the same-different task?} {Can deep convolutional neural networks support relational reasoning in the same-different task?}{\BBCQ}
\newblock
\APACjournalVolNumPages{Journal of Vision}{22}{10}{11--11}.
\PrintBackRefs{\CurrentBib}

\bibitem [\protect \citeauthoryear {%
Rumelhart%
\ \BBA {} McClelland%
}{%
Rumelhart%
\ \BBA {} McClelland%
}{%
{\protect \APACyear {1986}}%
}]{%
rumelhart1986learning}
\APACinsertmetastar {%
rumelhart1986learning}%
\begin{APACrefauthors}%
Rumelhart, D\BPBI E.%
\BCBT {}\ \BBA {} McClelland, J\BPBI L.%
\end{APACrefauthors}%
\unskip\
\newblock
\APACrefYearMonthDay{1986}{}{}.
\newblock
{\BBOQ}\APACrefatitle {On learning the past tenses of {English} verbs} {On learning the past tenses of {English} verbs}.{\BBCQ}
\newblock
\APACjournalVolNumPages{Psycholinguistics: Critical Concepts in Psychology}{4}{}{216--271}.
\PrintBackRefs{\CurrentBib}

\bibitem [\protect \citeauthoryear {%
Schmidhuber%
}{%
Schmidhuber%
}{%
{\protect \APACyear {1987}}%
}]{%
schmidhuber1987evolutionary}
\APACinsertmetastar {%
schmidhuber1987evolutionary}%
\begin{APACrefauthors}%
Schmidhuber, J.%
\end{APACrefauthors}%
\unskip\
\newblock
\APACrefYear{1987}.
\newblock
\APACrefbtitle {Evolutionary principles in self-referential learning} {Evolutionary principles in self-referential learning}.
\newblock
\BUPhD, Institut f{\"u}r Informatik, Technische Universit{\"a}t M{\"u}nchen.
\PrintBackRefs{\CurrentBib}

\bibitem [\protect \citeauthoryear {%
Shrager%
\ \BBA {} Langley%
}{%
Shrager%
\ \BBA {} Langley%
}{%
{\protect \APACyear {1990}}%
}]{%
ShragerLangley1990a}
\APACinsertmetastar {%
ShragerLangley1990a}%
\begin{APACrefauthors}%
Shrager, J.%
\BCBT {}\ \BBA {} Langley, P.%
\end{APACrefauthors}%
\ (\BEDS).
\unskip\
\newblock
\APACrefYear{1990}.
\newblock
\APACrefbtitle {Computational models of scientific discovery and theory formation} {Computational models of scientific discovery and theory formation}.
\newblock
\APACaddressPublisher{San Mateo, CA}{Morgan Kaufmann}.
\PrintBackRefs{\CurrentBib}

\bibitem [\protect \citeauthoryear {%
Spelke%
\ \BBA {} Kinzler%
}{%
Spelke%
\ \BBA {} Kinzler%
}{%
{\protect \APACyear {2007}}%
}]{%
spelke2007core}
\APACinsertmetastar {%
spelke2007core}%
\begin{APACrefauthors}%
Spelke, E\BPBI S.%
\BCBT {}\ \BBA {} Kinzler, K\BPBI D.%
\end{APACrefauthors}%
\unskip\
\newblock
\APACrefYearMonthDay{2007}{}{}.
\newblock
{\BBOQ}\APACrefatitle {Core knowledge} {Core knowledge}.{\BBCQ}
\newblock
\APACjournalVolNumPages{Developmental Science}{10}{1}{89--96}.
\PrintBackRefs{\CurrentBib}

\bibitem [\protect \citeauthoryear {%
Stabinger%
, Rodr{\'\i}guez-S{\'a}nchez%
\BCBL {}\ \BBA {} Piater%
}{%
Stabinger%
\ \protect \BOthers {.}}{%
{\protect \APACyear {2016}}%
}]{%
stabinger201625}
\APACinsertmetastar {%
stabinger201625}%
\begin{APACrefauthors}%
Stabinger, S.%
, Rodr{\'\i}guez-S{\'a}nchez, A.%
\BCBL {}\ \BBA {} Piater, J.%
\end{APACrefauthors}%
\unskip\
\newblock
\APACrefYearMonthDay{2016}{}{}.
\newblock
{\BBOQ}\APACrefatitle {25 years of {CNN}s: Can we compare to human abstraction capabilities?} {25 years of {CNN}s: Can we compare to human abstraction capabilities?}{\BBCQ}
\newblock
\BIn{} \APACrefbtitle {International {C}onference on {A}rtificial {N}eural {N}etworks ({ICANN})} {International {C}onference on {A}rtificial {N}eural {N}etworks ({ICANN})}\ (\BPGS\ 380--387).
\PrintBackRefs{\CurrentBib}

\bibitem [\protect \citeauthoryear {%
Tartaglini%
\ \protect \BOthers {.}}{%
Tartaglini%
\ \protect \BOthers {.}}{%
{\protect \APACyear {2023}}%
}]{%
tartaglini2023deep}
\APACinsertmetastar {%
tartaglini2023deep}%
\begin{APACrefauthors}%
Tartaglini, A\BPBI R.%
, Feucht, S.%
, Lepori, M\BPBI A.%
, Vong, W\BPBI K.%
, Lovering, C.%
, Lake, B\BPBI M.%
\BCBL {}\ \BBA {} Pavlick, E.%
\end{APACrefauthors}%
\unskip\
\newblock
\APACrefYearMonthDay{2023}{}{}.
\newblock
{\BBOQ}\APACrefatitle {Deep neural networks can learn generalizable same-different visual relations} {Deep neural networks can learn generalizable same-different visual relations}.{\BBCQ}
\newblock
\APACjournalVolNumPages{arXiv preprint arXiv:2310.09612}{}{}{}.
\PrintBackRefs{\CurrentBib}

\bibitem [\protect \citeauthoryear {%
Wang%
, Cao%
, De~Melo%
\BCBL {}\ \BBA {} Liu%
}{%
Wang%
\ \protect \BOthers {.}}{%
{\protect \APACyear {2016}}%
}]{%
wang2016relation}
\APACinsertmetastar {%
wang2016relation}%
\begin{APACrefauthors}%
Wang, L.%
, Cao, Z.%
, De~Melo, G.%
\BCBL {}\ \BBA {} Liu, Z.%
\end{APACrefauthors}%
\unskip\
\newblock
\APACrefYearMonthDay{2016}{}{}.
\newblock
{\BBOQ}\APACrefatitle {Relation classification via multi-level attention {CNN}s} {Relation classification via multi-level attention {CNN}s}.{\BBCQ}
\newblock
\BIn{} \APACrefbtitle {Proceedings of the 54th {A}nnual {M}eeting of the {A}ssociation for {C}omputational {L}inguistics} {Proceedings of the 54th {A}nnual {M}eeting of the {A}ssociation for {C}omputational {L}inguistics}\ (\BPGS\ 1298--1307).
\PrintBackRefs{\CurrentBib}

\bibitem [\protect \citeauthoryear {%
Webb%
\ \protect \BOthers {.}}{%
Webb%
\ \protect \BOthers {.}}{%
{\protect \APACyear {2024}}%
}]{%
webb2024relational}
\APACinsertmetastar {%
webb2024relational}%
\begin{APACrefauthors}%
Webb, T\BPBI W.%
, Frankland, S\BPBI M.%
, Altabaa, A.%
, Segert, S.%
, Krishnamurthy, K.%
, Campbell, D.%
\BDBL {}Cohen, J\BPBI D.%
\end{APACrefauthors}%
\unskip\
\newblock
\APACrefYearMonthDay{2024}{}{}.
\newblock
{\BBOQ}\APACrefatitle {The relational bottleneck as an inductive bias for efficient abstraction} {The relational bottleneck as an inductive bias for efficient abstraction}.{\BBCQ}
\newblock
\APACjournalVolNumPages{Trends in Cognitive Sciences}{}{}{}.
\PrintBackRefs{\CurrentBib}

\bibitem [\protect \citeauthoryear {%
Webb%
, Sinha%
\BCBL {}\ \BBA {} Cohen%
}{%
Webb%
\ \protect \BOthers {.}}{%
{\protect \APACyear {2021}}%
}]{%
webb2021emergent}
\APACinsertmetastar {%
webb2021emergent}%
\begin{APACrefauthors}%
Webb, T\BPBI W.%
, Sinha, I.%
\BCBL {}\ \BBA {} Cohen, J.%
\end{APACrefauthors}%
\unskip\
\newblock
\APACrefYearMonthDay{2021}{}{}.
\newblock
{\BBOQ}\APACrefatitle {Emergent Symbols through Binding in External Memory} {Emergent symbols through binding in external memory}.{\BBCQ}
\newblock
\BIn{} \APACrefbtitle {International {C}onference on {L}earning {R}epresentations.} {International {C}onference on {L}earning {R}epresentations.}
\PrintBackRefs{\CurrentBib}

\bibitem [\protect \citeauthoryear {%
Yamins%
\ \protect \BOthers {.}}{%
Yamins%
\ \protect \BOthers {.}}{%
{\protect \APACyear {2014}}%
}]{%
yamins2014performance}
\APACinsertmetastar {%
yamins2014performance}%
\begin{APACrefauthors}%
Yamins, D\BPBI L.%
, Hong, H.%
, Cadieu, C\BPBI F.%
, Solomon, E\BPBI A.%
, Seibert, D.%
\BCBL {}\ \BBA {} DiCarlo, J\BPBI J.%
\end{APACrefauthors}%
\unskip\
\newblock
\APACrefYearMonthDay{2014}{}{}.
\newblock
{\BBOQ}\APACrefatitle {Performance-optimized hierarchical models predict neural responses in higher visual cortex} {Performance-optimized hierarchical models predict neural responses in higher visual cortex}.{\BBCQ}
\newblock
\APACjournalVolNumPages{Proceedings of the National Academy of Sciences}{111}{23}{8619--8624}.
\PrintBackRefs{\CurrentBib}

\end{thebibliography}


\clearpage
\appendix
\section{Experimental Details}
Following Kim et al. 2018, we instantiated the CNN architectures as follows: 
\begin{itemize}
\item 2-Layer CNN (referred to as Conv2): two convolutional layers with 6, $2 \times 2$ convolutional filters in the first layer 
 \item 4-layer CNN (referred to as Conv4): four convolutional layers with 12, $4 \times 4$ filters in the first layer
\item 6-layer CNN (referred to as Conv6): six convolutional layers with 18, $6 ]times 6$ filters in the first layer. 

\end{itemize}

For all networks, in all subsequent convolutional layers, filter size is fixed at $2 \times 2$ with the number of filters doubling every layer. All convolutional layers had strides of 1 and used rectified linear (ReLU) activations. Pooling layers were placed after every convolutional layer, with pooling kernels of size $3 \times 3$ and strides of 2. On top of the retinotopic layers, all nine CNNs had three fully connected layers with 1024 hidden units in each layer and dropout with $p=0.5$, followed by a two-dimensional classification layer. All CNNs were trained on all problems. Network parameters were initialized using Xavier initialization and were trained using the Adaptive Moment Estimation (Adam) optimizer with a base learning rate of $\alpha = 1e^{-3}$. All experiments were run using PyTorch and were run until convergence with patience set to stop after failing to improve by over 1 percent over the course of 10 epochs.

\begin{table}[H]
\centering
\caption{In-distribution generalization performance of conventional learning with Adam and meta-learning with MAML over the same-different data from Puebla and Bowers (Figure 3). \normalsize Results show mean accuracy (\%) ± standard deviation over 10 random seeds for each CNN architecture on the tasks from \citep{puebla2022can}. Chance performance is 50\%.}
\resizebox{\textwidth}{!}{%
\begin{tabular}{|l|c|c|c||c|c|c|}
\hline
& \multicolumn{3}{c||}{\textbf{Conventional Learning (Acc \%)}} & \multicolumn{3}{c|}{\textbf{Meta-Learning (Acc \%)}} \\
\hline
\textbf{Task} & \textbf{2-Layer} & \textbf{4-Layer} & \textbf{6-Layer} & \textbf{2-Layer} & \textbf{4-Layer} & \textbf{6-Layer} \\
\hline
regular & $50.0 \pm 0.0$ & $50.0 \pm 0.0$ & $50.0 \pm 0.0$ & $52.2 \pm 5.0$ & $85.8 \pm 9.9$ & $98.5 \pm 0.7$ \\
lines & $79.5 \pm 24.1$ & $50.0 \pm 0.0$ & $50.0 \pm 0.0$ & $78.6 \pm 8.0$ & $94.8 \pm 14.9$ & $100.0 \pm 0.1$ \\
open & $51.4 \pm 4.3$ & $50.0 \pm 0.0$ & $50.0 \pm 0.0$ & $48.8 \pm 6.0$ & $85.2 \pm 10.3$ & $97.5 \pm 0.8$ \\
wider\_line & $50.0 \pm 0.0$ & $50.0 \pm 0.0$ & $50.0 \pm 0.0$ & $52.2 \pm 5.0$ & $84.2 \pm 10.4$ & $97.8 \pm 0.7$ \\
scrambled & $99.1 \pm 0.6$ & $50.0 \pm 0.0$ & $50.0 \pm 0.0$ & $82.0 \pm 7.0$ & $94.4 \pm 14.8$ & $99.9 \pm 0.1$ \\
random\_color & $55.0 \pm 0.0$ & $55.0 \pm 0.0$ & $55.0 \pm 0.0$ & $49.8 \pm 4.0$ & $86.3 \pm 7.2$ & $95.0 \pm 0.7$ \\
arrows & $50.0 \pm 0.0$ & $50.0 \pm 0.0$ & $50.0 \pm 0.0$ & $48.6 \pm 6.0$ & $69.0 \pm 11.6$ & $93.5 \pm 1.8$ \\
irregular & $50.0 \pm 0.0$ & $54.2 \pm 12.5$ & $50.0 \pm 0.0$ & $45.4 \pm 5.0$ & $83.0 \pm 12.3$ & $99.2 \pm 0.4$ \\
filled & $52.4 \pm 7.2$ & $50.0 \pm 0.0$ & $50.0 \pm 0.0$ & $54.0 \pm 6.0$ & $84.2 \pm 13.4$ & $97.9 \pm 0.6$ \\
original & $50.0 \pm 0.0$ & $50.0 \pm 0.0$ & $50.0 \pm 0.0$ & $51.0 \pm 5.0$ & $84.0 \pm 10.9$ & $97.0 \pm 1.3$ \\
\hline
\textbf{Average} & $58.7 \pm 16.0$ & $50.9 \pm 1.7$ & $50.5 \pm 1.6$ & $56.3 \pm 12.6$ & $85.1 \pm 7.2$ & $97.6 \pm 2.0$ \\
\hline
\end{tabular}%
}%
\centering
\vspace{5pt}

\end{table}

\vspace{10pt}


\subsection{Episode Visualizations}
Here we show how meta-learning episodes are structured and presented to the model at train/test time (varying between 4, 6, 8, and 10 labeled examples and fixing queries at 2 examples). The full gradient calculations over the labeled and query sets are given in Appendix B. 
\begin{figure}[H]
    \centering
    \includegraphics[width=0.8\columnwidth]{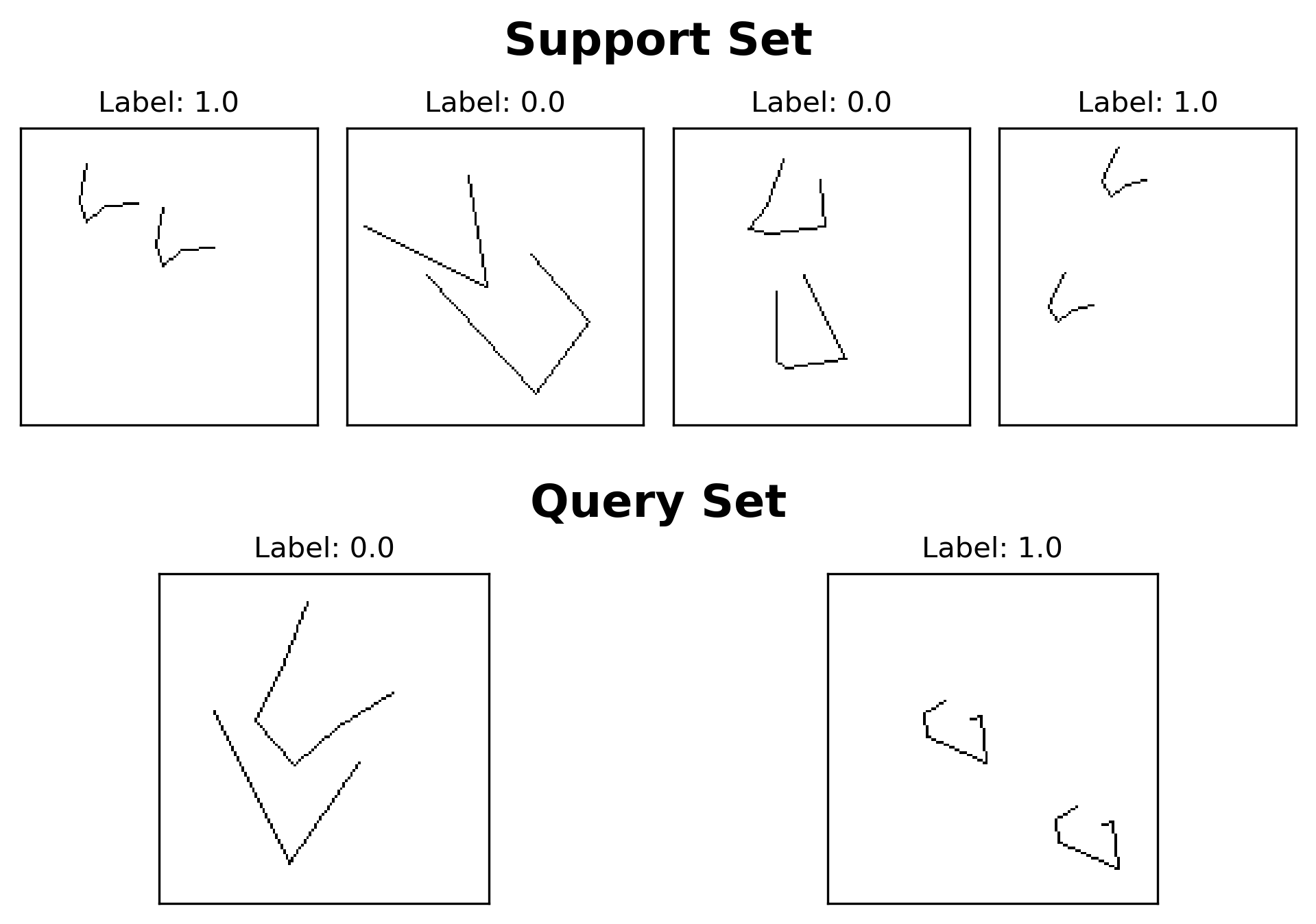}
    \caption{A single episode of the ``open'' task from Puebla and Bowers (2022) as presented to the model at training and test time. Note: this particular example has 4 labeled examples and 2 queries, but we randomly intermix the sizes of the sets of labeled examples to promote better generalization.
    }
    \label{fig:open_episode}
\end{figure}

\begin{figure}[H]
    \centering
    \includegraphics[width=0.9\columnwidth]{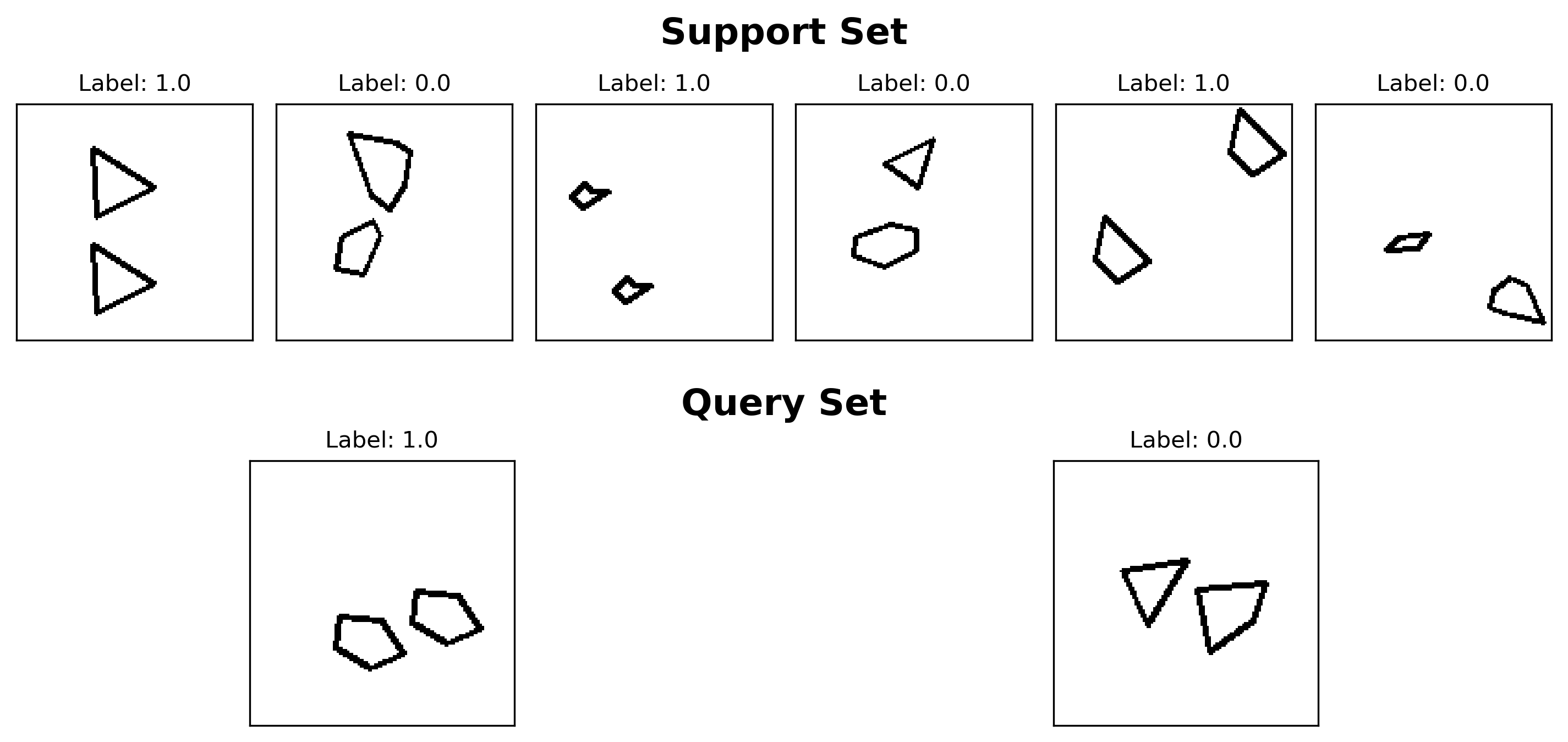}
    \caption{An episode from the ``wider line'' task with 6 labeled examples.}
    \label{fig:wider_line_episode}
\end{figure}

\begin{figure}[H]
    \centering
    \includegraphics[width=0.9\columnwidth]{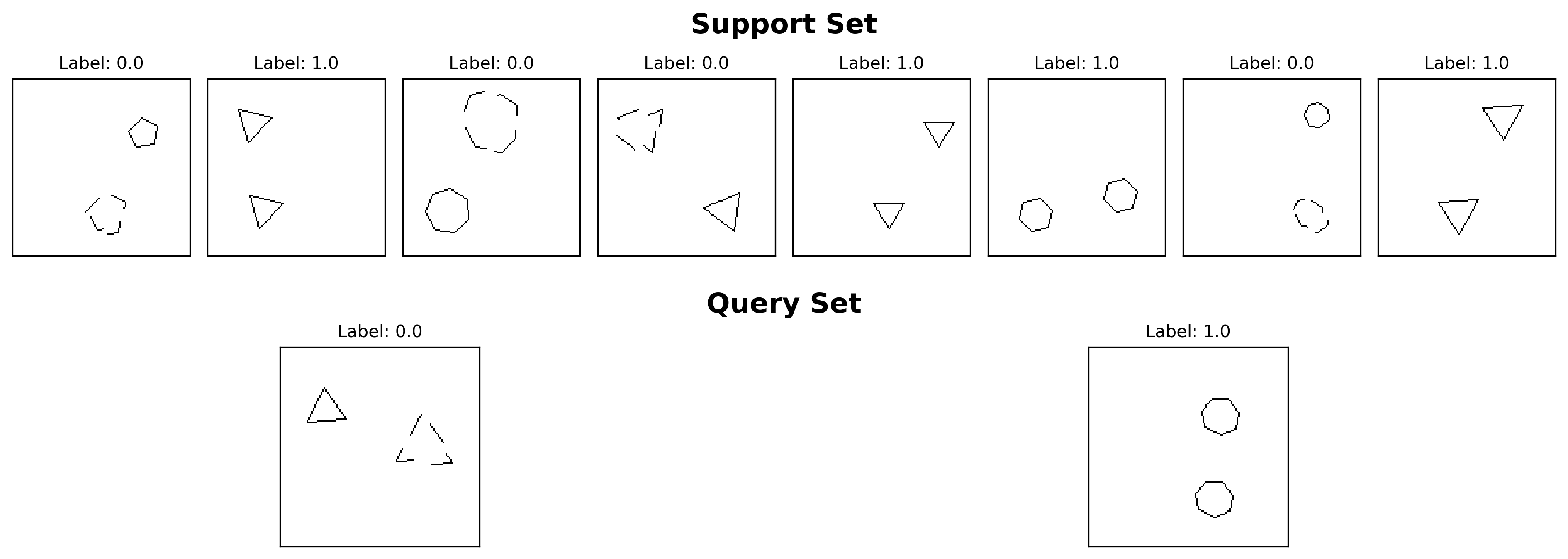}
    \caption{An episode from the ``scrambled'' task with  8 labeled examples.}
    \label{fig:scrambled_episode}
\end{figure}

\begin{figure}[H]
    \centering
    \includegraphics[width=0.9\columnwidth]{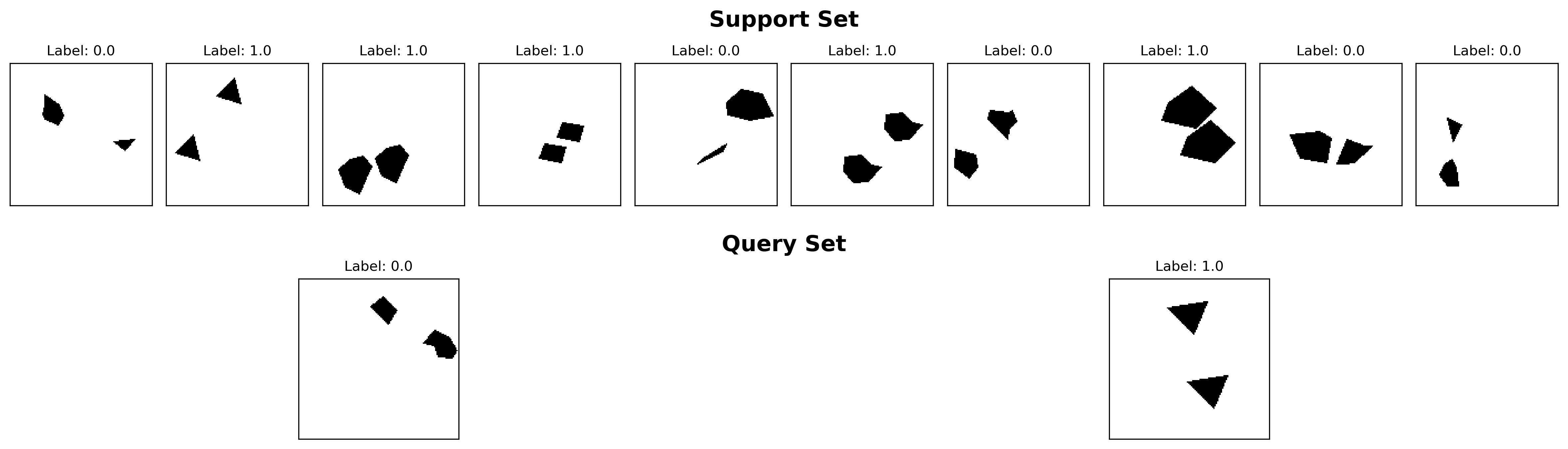}
    \caption{An episode from the ``filled'' task with 10 labeled examples.}
    \label{fig:filled_episode}
\end{figure}

\section{Role of the Meta-Gradient in MAML}
\label{app:maml_derivation}

Below we show how the second derivative emerges in second-order Model-Agnostic Meta-Learning (MAML) used throughout this paper. 
For a given task \(\tau\) -- e.g.\ one of the ten
shape families in Figure \ref{fig:augmented} -- each episode provides  

\begin{itemize}
    \item a set of \textbf{labeled examples}, on which we compute the \emph{inner-loop}
          loss \(\mathcal{L}^{(0)}_{\tau}\) (Eq.~\ref{eq:inner_steps}), and  
    \item a set of \textbf{queries}, on which we compute the
          \emph{outer-loop} loss \(\mathcal{L}^{(1)}_{\tau}\)
          (Eq.~\eqref{1}).
\end{itemize}

\noindent {\bf Inner loop (fast adaptation)}
Starting from the meta parameters \(\theta_{\text{meta}}\equiv\theta_{0}\),
we perform \(k\) gradient steps on the labeled examples:

\begin{align}
\theta_{0} &= \theta_{\text{meta}}\\
\theta_{1} &= \theta_{0}-\alpha\nabla_{\theta}\mathcal{L}^{(0)}_{\tau}(\theta_{0})\\
\theta_{2} &= \theta_{1}-\alpha\nabla_{\theta}\mathcal{L}^{(0)}_{\tau}(\theta_{1})\\
           &\;\;\vdots \nonumber \\
\theta_{k} &= \theta_{k-1}-\alpha\nabla_{\theta}\mathcal{L}^{(0)}_{\tau}(\theta_{k-1}). \label{eq:inner_steps}
\end{align}

\noindent {\bf Outer loop (meta-update)}
We then update the meta parameters using the query loss:
\[
\theta_{\text{meta}}\;\leftarrow\;
\theta_{\text{meta}}-\beta\,g_{\text{MAML}},
\]
where the exact MAML gradient is: 
\\
\begin{align}
g_{\text{MAML}}
  &= \nabla_{\theta}\mathcal{L}^{(1)}_{\tau}(\theta_{k})
     \label{eq:maml_grad_start}\\[6pt]
  &= \nabla_{\theta_{k}}\mathcal{L}^{(1)}_{\tau}(\theta_{k})
     \prod_{i=1}^{k}\frac{\partial\theta_{i}}{\partial\theta_{i-1}}
     \quad\text{\footnotesize(chain rule)}
     \label{eq:maml_grad_chain}\\[6pt]
  &= \nabla_{\theta_{k}}\mathcal{L}^{(1)}_{\tau}(\theta_{k})
     \prod_{i=1}^{k}\nabla_{\theta_{i-1}}
        \bigl(\theta_{i-1}-\alpha\nabla_{\theta}
        \mathcal{L}^{(0)}_{\tau}(\theta_{i-1})\bigr)
     \label{eq:maml_grad_expand}\\[6pt]
  &= \nabla_{\theta_{k}}\mathcal{L}^{(1)}_{\tau}(\theta_{k})
     \prod_{i=1}^{k}\Bigl(I-\alpha\nabla_{\theta_{i-1}}
        \nabla_{\theta}\mathcal{L}^{(0)}_{\tau}(\theta_{i-1})\Bigr)
     \label{1}
\end{align}

In our experiments each task $\tau$ corresponds to a specific shape
category (\textit{regular polygons, arrows, lines, etc.}). The loss on the labeled examples
$\mathcal{L}^{(0)}_{\tau}$ is therefore computed on a small batch (4--10 images)
of same-different examples drawn \textit{within} that category, while the query
loss $\mathcal{L}^{(1)}_{\tau}$ is evaluated on two different held-out examples
from the same category. The meta-gradient above thus aligns the initial weights
to learn each new same-different category quickly.

\begin{figure}[b!]
    \centering
    \includegraphics[width=0.85\columnwidth]{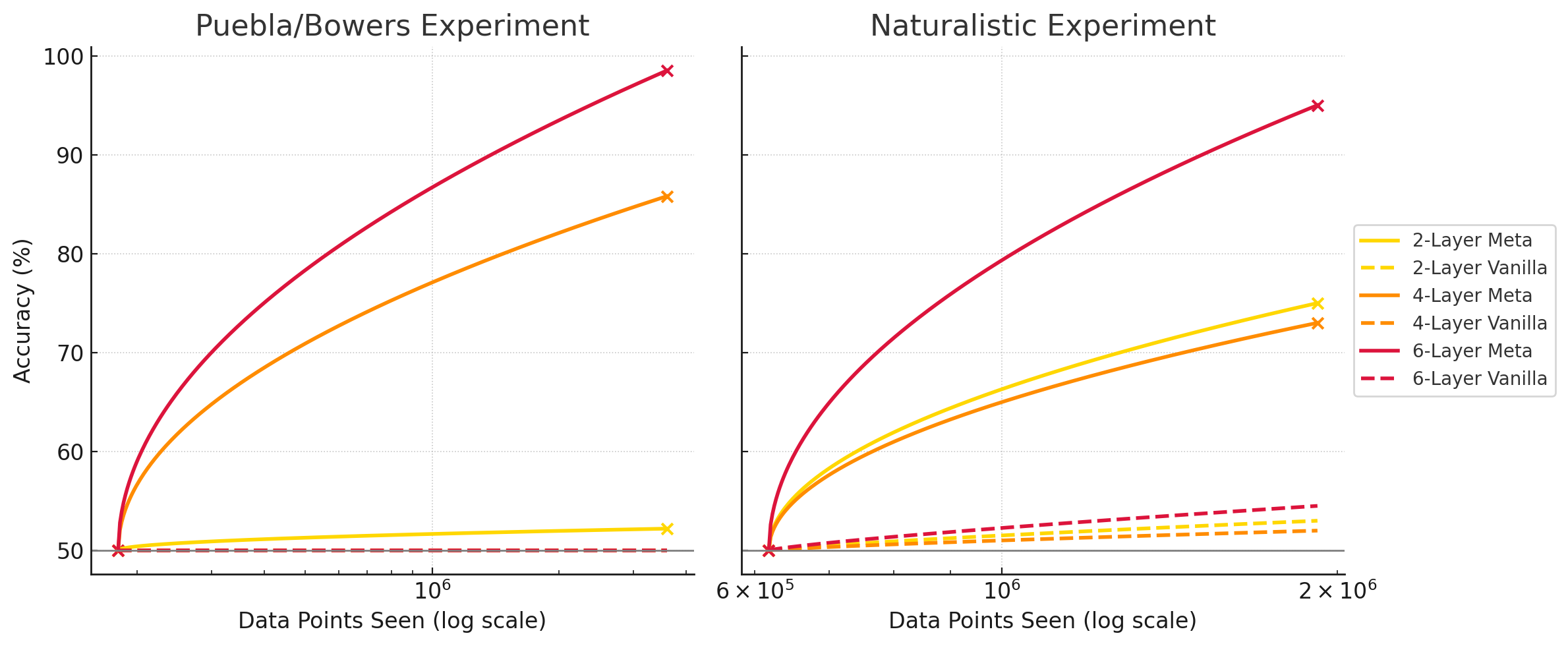}
    \caption{Train-time sample efficiency of conventional (Adam) training versus second-order MAML across 2-, 4-, and 6-layer architectures in both synthetic and naturalistic experiments.}
    \label{fig:sample-efficiency}
\end{figure}

The final product in Eq.~(1) contains \emph{second-order} terms (Hessians of the
support loss) that are expensive to compute upfront (during training) but render
test-time compute much more efficient and effective, as we document in
Figure~\ref{fig:sample-efficiency}. Discarding these terms yields the commonly
used \emph{first-order MAML} (FOMAML) approximation:
\[
g_{\text{FOMAML}}=\nabla_{\theta_k}\mathcal{L}^{(1)}_{\tau}(\theta_k).
\]
Although train-time efficiency gains are saved by this approximation, we ran
additional experiments to confirm that the second-order gradient of MAML is
causally implicated in model performance on our same-different tasks. Ablating
the second-order gradient downgrades model performance back to chance level
(Table~\ref{tab:fomaml_vs_maml_fullwidth}).


\begin{table}[H]
\centering
\caption{In-distribution test accuracy (\%) for first-order MAML versus second-order MAML. C2/C4/C6 denote 2-, 4-, and 6-layer CNNs. Chance level is 50\%.}
\small
\setlength{\tabcolsep}{6pt}        
\renewcommand{\arraystretch}{1.05} 
\begin{tabular*}{\textwidth}{@{\extracolsep{\fill}} l|ccc||ccc @{}}
\toprule
 & \multicolumn{3}{c||}{\textbf{First-order MAML}} & \multicolumn{3}{c}{\textbf{Second-order MAML}}\\
\cmidrule(lr){2-4}\cmidrule(lr){5-7}
\textbf{Task} & \textbf{C2} & \textbf{C4} & \textbf{C6} & \textbf{C2} & \textbf{C4} & \textbf{C6}\\
\midrule
regular        & \(0.50\!\pm\!0.01\) & \(0.51\!\pm\!0.04\) & \(0.52\!\pm\!0.05\) & \(52.2\!\pm\!5.0\) & \(85.8\!\pm\!9.9\)  & \(98.5\!\pm\!0.7\) \\
lines          & \(0.50\!\pm\!0.01\) & \(0.50\!\pm\!0.01\) & \(0.56\!\pm\!0.15\) & \(78.6\!\pm\!8.0\) & \(94.8\!\pm\!14.9\) & \(100.0\!\pm\!0.1\) \\
open           & \(0.50\!\pm\!0.01\) & \(0.51\!\pm\!0.05\) & \(0.52\!\pm\!0.05\) & \(48.8\!\pm\!6.0\) & \(85.2\!\pm\!10.3\) & \(97.5\!\pm\!0.8\) \\
wider\_line    & \(0.50\!\pm\!0.01\) & \(0.51\!\pm\!0.04\) & \(0.52\!\pm\!0.05\) & \(52.2\!\pm\!5.0\) & \(84.2\!\pm\!10.4\) & \(97.8\!\pm\!0.7\) \\
scrambled      & \(0.51\!\pm\!0.02\) & \(0.50\!\pm\!0.01\) & \(0.50\!\pm\!0.01\) & \(82.0\!\pm\!7.0\) & \(94.4\!\pm\!14.8\) & \(99.9\!\pm\!0.1\) \\
random\_color  & \(0.50\!\pm\!0.00\) & \(0.50\!\pm\!0.00\) & \(0.50\!\pm\!0.00\) & \(49.8\!\pm\!4.0\) & \(86.3\!\pm\!7.2\)  & \(95.0\!\pm\!0.7\) \\
arrows         & \(0.50\!\pm\!0.01\) & \(0.50\!\pm\!0.01\) & \(0.50\!\pm\!0.01\) & \(48.6\!\pm\!6.0\) & \(69.0\!\pm\!11.6\) & \(93.5\!\pm\!1.8\) \\
irregular      & \(0.50\!\pm\!0.01\) & \(0.51\!\pm\!0.05\) & \(0.52\!\pm\!0.06\) & \(45.4\!\pm\!5.0\) & \(83.0\!\pm\!12.3\) & \(99.2\!\pm\!0.4\) \\
filled         & \(0.50\!\pm\!0.01\) & \(0.51\!\pm\!0.05\) & \(0.52\!\pm\!0.04\) & \(54.0\!\pm\!6.0\) & \(84.2\!\pm\!13.4\) & \(97.9\!\pm\!0.6\) \\
original       & \(0.50\!\pm\!0.01\) & \(0.51\!\pm\!0.04\) & \(0.51\!\pm\!0.04\) & \(51.0\!\pm\!5.0\) & \(84.0\!\pm\!10.9\) & \(97.0\!\pm\!1.3\) \\
\midrule
\textbf{Average} & \(0.50\!\pm\!0.01\) & \(0.51\!\pm\!0.04\) & \(0.52\!\pm\!0.06\) & \(56.3\!\pm\!12.6\) & \(85.1\!\pm\!7.2\)  & \(97.6\!\pm\!2.0\) \\
\bottomrule
\end{tabular*}
\vspace{5pt}

\label{tab:fomaml_vs_maml_fullwidth}
\end{table}

\section{Sample Efficiency}
\label{app:maml_derivation}

We are interested in the number of samples required for a model to be able to start generalizing effectively at test time. To test the samply efficiencies of MAML versus conventional training in this domain, we ran a separate sample efficiency test measuring the accuracy of models as a function of data points seen in both synthetic and naturalistic experiments.  
Figure 12 shows that across both naturalistic and synthetic experiments (left and right panels respectively), meta-learning generalizes much more  efficiently than the conventional comparisons. The decision then on whether or not to use a first-order approximation is a tradeoff between compute speed and test-time accuracy. If we wish to spend more resources at train-time to generalize more effectively at test-time, these results show that second-order MAML is more efficient. If we wish to spend minimally at train-time and compromise test-time accuracy, we might settle for a first-order approximation. The fact that the second-order gradient is causally implicated in model performance tells us that the loss landscape encoded by the same-different function might not be totally convex, which could explain the difficulties encountered by conventional and first-order training methods. 
\end{document}